\documentclass[twoside]{article}
\usepackage{ecj,palatino,epsfig,latexsym,natbib}
\pdfoutput=1
\usepackage{hyperref}

\usepackage{lmodern}
\usepackage{amssymb,amsmath,amstext}
\usepackage[ruled,vlined]{algorithm2e}
\usepackage{graphicx} 
\usepackage{longtable,booktabs}
\usepackage[utf8]{inputenc}
\usepackage{array,multirow}   
\usepackage{subcaption}
\usepackage{lscape} 

\newsavebox\CBox
\def\textBF#1{\sbox\CBox{#1}\resizebox{\wd\CBox}{\ht\CBox}{\textbf{#1}}}

\begin{document}


  \title{\bf Shape-constrained Symbolic Regression -- Improving Extrapolation with Prior Knowledge}

\author{
          \name{\bf G. Kronberger} \hfill \addr{gabriel.kronberger@fh-ooe.at} \\
          \addr{Josef Ressel Center for Symbolic Regression, University
                of Applied Sciences Upper Austria, Softwarepark 11, 4232 Hagenberg, Austria}
\AND
          \name{\bf F. O. de Franca} \hfill \addr{folivetti@ufabc.edu.br}\\
          \addr{Center for Mathematics, Computation and Cognition (CMCC), Heuristics, Analysis and Learning Laboratory (HAL), Federal University of ABC,
          Santo Andre, Brazil}
\AND
         \name{\bf B. Burlacu} \and {\bf C. Haider} \and {\bf M. Kommenda} \hfill \addr{}\\
         \addr{Josef Ressel Center for Symbolic Regression, University
                of Applied Sciences Upper Austria, Softwarepark 11, 4232 Hagenberg, Austria}
  }

\maketitle 
  
\begin{abstract}
We investigate the addition of constraints on the function image and
its derivatives for the incorporation of prior knowledge in symbolic
regression. The approach is called shape-constrained symbolic
regression and allows us to enforce e.g.  monotonicity of the function
over selected inputs. The aim is to find models which conform to
expected behaviour and which have improved extrapolation capabilities.
We demonstrate the feasibility of the idea and propose and compare two
evolutionary algorithms for shape-constrained symbolic regression: i)
an extension of tree-based genetic programming which discards
infeasible solutions in the selection step, and ii) a two population
evolutionary algorithm that separates the feasible from the infeasible
solutions. In both algorithms we use interval arithmetic to
approximate bounds for models and their partial derivatives.  The
algorithms are tested on a set of 19 synthetic and four real-world
regression problems. Both algorithms are able to identify models which
conform to shape constraints which is not the case for the unmodified
symbolic regression algorithms. However, the predictive accuracy of
models with constraints is worse on the training set and the test
set. Shape-constrained polynomial regression produces the best results
for the test set but also significantly larger models.\footnote{This work has been accepted to be published by Evolutionary Computation, MIT Press}
\end{abstract}


\section{Introduction and Motivation}\label{introduction-and-motivation}

Dynamical systems and processes in critical application areas such as
control engineering require trustworthy, robust and reliable models
that strictly conform to behavioral expectations according to
domain-specific knowledge, as well as safety and performance criteria.
However, such models cannot always be derived from first principles
and theoretical considerations alone. Thus, they must be determined
empirically.
From this perspective, models used in system identification can be categorized as: 
\begin{itemize}
    \item \emph{White-box}. Derived from first principles, explicitly include
        domain knowledge, can be valid for a wide range of inputs even beyond
        available observations and allow to make far-reaching predictions
        (extrapolation).  
    \item \emph{Gray-box}. Derived from data, with known internals, open for
        further inspection. Can be validated against domain knowledge. Only
        valid for inputs which are similar to observations that are used for
        model fitting.  
    \item \emph{Black-box}. Derived from data, with unknown internals (e.g.,
        neural networks), difficult or impossible to inspect and/or validate. 
\end{itemize}

In-between the two extremes there is a whole spectrum of gray-box models which
combine aspects of both extremes \citep{Ljung2010}.  The proponents of
``explainable AI'' propose to apply models on the white end of this spectrum
either directly \citep{Rudin2019} or as an approximation of black-box models in
AI technologies.

Empirical, data-driven models are required in scenarios where first principles
models are infeasible due to engineering or financial considerations. Ensuring
that empirical models conform to expected system behavior, correctly reflect
physical principles, and can extrapolate well on unseen data remains an open
challenge in this area. 

Symbolic regression (SR) is especially interesting in this context because SR
models are closed-form expressions which are structurally similar to white-box
models derived from first principles.  The aim in SR is to identify the best
function form and parameters for a given data set using common mathematical
operators and functions which serve as building blocks for the function form
\citep{koza1992genetic}. This is in contrast to other forms of regression
analysis where the function form is pre-specified and only the numerical
coefficients are optimized through model fitting.

It should be noted however that SR is fundamentally a purely data-driven
approach. It does not require prior information about the modelled process or
system and leads to gray-box models which can be hard to understand. SR
algorithms therefore favor shorter expressions and use simplification to
facilitate detailed analysis of their models. 

The most well-known solution methods for SR includes
tree-based genetic programming (GP) variants~\citep{koza1992genetic},
linear GP~\citep{oltean2003comparison} including
Pareto-GP~\citep{smits2005pareto}, Cartesian GP~\citep{miller2008cartesian},
grammatical evolution (GE)~\citep{o2001grammatical}, and gene expression
programming (GEP)~\citep{ferreira2001gene}. More recent developments of 
SR include Geometric Semantic GP~\citep{Moraglio2012, Pawlak2018, Ruberto2020}, 
Multiple-regression GP~\citep{Arnaldo2014}, and a memetic evolutionary algorithm using a novel fractional representation for SR~\citep{Sun2019}.  There are also several
non-evolutionary algorithms for SR including fast function extraction
(FFX)~\citep{mcconaghy2011ffx}, SymTree~\citep{de2018greedy}, prioritized
grammar enumeration (PGE)~\citep{worm2013prioritized}. Recently several neural network
architectures have been used for symbolic regression ~\citep{sahoo2018learning, Kim2019IntegrationON, petersen2019deep, udrescu2020ai}.

This contribution introduces an extension of SR, which allows to include
vague prior knowledge via constraints on the shape of SR model image and
derivatives. We therefore call this approach shape-constrained SR as it is a specific form of shape-constrained regression.  The
main motivation is, that even in data-based modelling tasks, partial knowledge
about the modelled system or process is often available, and can be helpful to
produce better models which conform to expected behaviour and might improve
extrapolation.
Such information could be used for:
\begin{enumerate}
\def\labelenumi{\arabic{enumi}.}
\item
  data augmentation to decrease the effects of noisy measurements or
  for detecting implausible outliers
\item
  fitting the model in input space regions where observations are scarce or unavailable
\item
  distinguishing between correlations in observation data and causal
  dependencies
\item
  increasing the efficiency of the statistical modelling technique.
\end{enumerate}

\subsection{Problem Statement}\label{problem-statement}

We focus on the integration of prior knowledge in SR
solvers in order to find expressions compatible with the expected
behavior of the system.

Our assumptions are that we have data in the form of measurements as
well as side information about the general behaviour of the modelled
system or process that however is insufficient to formulate an explicit
closed-form expression for the model. We additionally assume that the
information can be expressed in the form of constraints on the
function output as well as its partial derivatives.

The task is therefore to 
find a closed-form expression 
which fits the data, possibly with a small error, and conforms to our
expectation of the general behaviour of the system. In particular, the
model must not violate shape constraints when interpolating and
extrapolating within a bounded input space.

We investigate the hypotheses that (i) interval arithmetic can be used to
approximate bounds for the image of SR models and their partial
derivatives and therefore reject non-conforming models
within SR solvers, and (ii) that the predictive accuracy of SR models can be improved by using side information in the form of shape constraints.


\section{Related Work}\label{prior-work}
The issue of knowledge
integration into machine learning methods including genetic
programming has been discussed before.

Closely related to this work is the work of \cite{Bladek} who have
discussed using formal constraints in combination with symbolic
regression to include domain knowledge about the monotonicity,
convexity or symmetry of functions. Their approach called
Counterexample-Driven GP utilizes satisfiability modulo theories (SMT)
solvers to check whether candidate SR models satisfy the constraints
and to extend the training set with counter examples. \cite{Bladek}
observe that is very unlikely for GP to synthesize models which
conform to constraints. The difference to this work is that we do not
support symmetry constraints and instead of a SAT solver we use
interval arithmetic to calculate bounds for the function image and its
partial derivatives. We test our approach on a larger set of harder
benchmark problems.

\cite{Versino2017data} have specifically studied the
generation of flow stress models for copper using GP with and without
knowledge integration. They have investigated four different
scenarios: i) a canonical GP solver that returns a Pareto front of 
  model quality and complexity, ii) knowledge integration by means of 
  transformed variables, samples weighting, introduction of artificial
  points, introduction of seed solutions, iii) search for the 
  derivative of the expression and then reconstructing the
  original equation by integration, iv) search derivatives but with
  additional artificial data points.

This work emphasized the importance of integrating prior knowledge in
the model fitting process. The authors observed that the canonical GP
solver presented unpredictable behavior w.r.t. the physical
constraints, thus the sampled data set alone was insufficient to guide
the solver into a model with both low error and physical
feasibility. In contrast to our approach which guarantees that
solutions conform to prior knowledge, all techniques tried in
\citep{Versino2017data} are optimistic and might produce infeasible
solutions. Additionally, adding artificial data points is subjective
and does not scale to high-dimensional problems.

\cite{schmidt2009incorporating} studied the influence of
integrating expert knowledge into the evolutionary search process
through a process called seeding. First the authors generated
approximation solutions by either solving a simpler problem or finding
an approximate solution to a more complex problem. These solutions are
then used during the seeding procedure by inserting approximations 
into the initial population, shuffled expressions and building blocks.
The authors found that seeding significantly improves the convergence
and fitness average performance when compared to no seeding. Among the
seeding procedures, the seeding of building blocks was the most successful
allowing faster convergence even in more complex problems. Again, seeding is
an optimistic approach and does not guarantee that the final SR
solution conforms to prior knowledge.

\cite{stewart2017label} investigated how to use prior
knowledge in artificial neural networks. Specifically, they incorporate physical
knowledge into motion detection and tracking and causal relationships in
object detection.

Even though the experiments were all small scaled, they showed
promising results and opened up the possibility of teaching an
artificial neural network by only describing the properties of the
approximation function. This idea is similar to the idea of
shape-constraints.

Recently an approach for knowledge integration for symbolic regression
with a similar motivation to this work has been described
in~\citep{li2019neural}. Instead of shape constraints the authors make
use of semantic priors in the form of leading powers for
symbolic expressions. They describe a Neural-Guided Monte Carlo Tree
Search (MCTS) algorithm to search of SR models which conform to prior
knowledge. The authors use a context-free grammar for
generating symbolic expressions and use MCTS for generating solutions whereby a neural network
predicts the next production rule given a sequence of
already applied rules and the leading power constraints.

The algorithm was compared against variations of MCTS and
a GP implementation using the DEAP framework~\citep{fortin2012deap}

The results show a superior performance of neural-guided MCTS with a
high success rate for easier instances but much lower rates on
harder instances which were however still higher than the success rate of GP.

One aim of our work is improving extrapolation of SR
solutions. Similarly to polynomial regression, SR models might produce
extreme outputs values when extrapolating. Interestingly,
extrapolation behaviour of SR models has been largely ignored in the
literature with only a few exceptions.

\cite{castillo2013symbolic}
investigated the difference in the response variation of
different models generated by GP to three different data sets. They have
shown that in some situations a GP algorithm can generate models with
different behavior while presenting a similar $R^2$ value. These
results motivate the need of a post analysis of the validity of the
model w.r.t. the system being studied and a verification of how well
such models extrapolate.

The extrapolation capability of GP is also studied
in~\citep{castillo2003methodology}.
In this work, the authors evaluate and compare the best
solution obtained by a GP algorithm to a linear model which uses the
transformed variables found by the GP model.

The experiments suggest that the GP models have good interpolation
capabilities but with a moderate extrapolation error. The proposed approach
presented good interpolation and extrapolation capabilities, suggesting
that at the very least, the GP models can guide the process of building
a new transformed space of variables.

Another relevant result was obtained by~\cite{kurse2012extrapolatable}
who used Eureqa (a commercial SR solver) to find
analytical functions that correctly modelled complex neuromuscular
systems. The experiments showed that the GP solver was capable of
extrapolating the data set much better than the traditional polynomial
regression, used to model such systems.

\cite{stewart2017label} trained a neural network by introducing
prior knowledge through a penalty term of the loss function 
such that the generated model is consistent to a physics behavior. Different
from our work, this generates a black box model of the data.
Finally, \cite{zhu2019physics} incorporate
the appropriate partial differential equations into the loss functions of a
neural network and, by doing so, they can train their model on unlabeled data, 
this approach requires more information than usually available when modelling
studied observations.

\section{Methods}\label{methods}
In this section we describe in detail the methods proposed in this paper to
integrate prior knowledge into SR solvers using interval
arithmetic and shape constraints.
In summary, we extend two GP variants to support shape-constrained SR:
a tree-based GP approach with optional memetic
improvement of models parameters and the Interaction-Transformation
Evolutionary Algorithm (ITEA)~\citep{de2019interaction}.  The extended algorithms are compared
against the original versions on a set of benchmark problems. For
the comparison we calculate the number of constraint violations and
the training as well as the test errors and qualitatively assess the
extrapolation behaviour.

\subsection{Shape-constrained Regression}\label{shape-constrained-regression}
Shape-constrained regression allows to fit a model whereby certain
characteristics of the model can be constrained. It is a general concept, that
encompasses a diverse set of methods including parametric and non-parametric,
as well as uni-variate and multi-variate methods. Examples include isotonic
regression~\citep{wright1980isotonic,Tibshirani2011}, monotonic lattice
regression~\citep{Gupta2016}, nonparametric shape-restricted
regression~\citep{guntuboyina2018nonparametric}, non-negative
splines~\citep{papp2014shape}, and shape-constrained polynomial
regression~\citep{Hall2018}.

In shape-constrained regression, the model is a function mapping
real-valued inputs to a real-valued output and the constraints refer
to the shape of the function. For instance if the function
$f(\mathbf{x})$ should be monotonically increasing over a given input
variable $x_1$ then we would introduce the monotonicity constraint
\begin{equation*}
  \frac{\partial f}{\partial x_1}(\mathbf{x}) \geq 0, \mathbf{x} \in \mathcal{S} \subseteq \mathbb{R}^d
\end{equation*}
Similarly, we could enforce concavity/convexity of a function by constraining
second order derivatives. Usually the input space for the model is limited for example to a
 $d$-dimensional box $\mathcal{S} = [\inf_1, \sup_1] \times [\inf_2,
  \sup_2] \times \dots \times [\inf_d, \sup_d] \subseteq \mathbb{R}^d$.

In general it is possible to include constraints for the model and its partial
derivatives of any order. However, in practice first and second order partial
derivatives are most relevant. The set of constraints $C$ contains expressions
that are derived from the model via one of the operators Op and linearly
transformed using a sign $s$ and threshold $c$. We use this representation of
shape constraints to simplify checking of constraints. All constraint
expressions $c_i\in C$ must not be positive for feasible solutions.
\begin{equation*}
  C = \left\{ \operatorname{s}\cdot \operatorname{Op}(f)(\mathbf{x}) - c \leq 0 \middle| \operatorname{Op} \in
  \left\{\operatorname{id}(f), \frac{\partial^n f}{\partial^n x_1}, \dots \frac{\partial^n f}{\partial^n x_d}\right\}, c \in \mathbb{R}, \operatorname{s} \in \{1, -1\}, n>0\right\}
\end{equation*}

It is important to mention that shape constraints limit the function
outputs and partial derivatives, but the optimal function which fits
the observed data still needs to be identified by the algorithm.
Shape-constrained regression is a general concept which is
applicable to different forms of regression analysis. For specific models 
like multi-variate linear regression or tree-based methods (e.g., XGBoost), it is easy
to integrate shape constraints. For other models, for instance, 
polynomial regression, it is harder to incorporate shape constraints
\citep{Hall2018, Ahmadi2019}.

Generally, one can distinguish
optimistic and pessimistic approaches to shape-constrained regression
\citep{Gupta2016}. Optimistic approaches check the constraints only
for a finite set of points from the input space and accept that the
identified model might violate the constraints for certain elements from
the input space. Pessimistic approaches calculate bounds for the
outputs of the model and its partial derivatives to guarantee that
identified solutions are in fact feasible. However, pessimistic
approaches might reject optimal models as a consequence of overly wide bounds.
In this paper, we decided to study a pessimistic approach and calculate bounds
for SR models and their derivatives using interval arithmetic.

\subsection{Interval Arithmetic}\label{interval-arithmetic}

Interval Arithmetic (IA) is a method for calculating output ranges for mathematical expressions~\citep{hickey2001interval}.
It has many
uses such as dealing with uncertainties stemming from inaccurate representations and verifying boundary
conditions. An interval is represented as $[a, b]$ with $a \leq b$,
named lower and upper endpoints, respectively.

If we have a function $f(x_1, x_2) = x_1 + x_2$, for example, and
knowing the intervals for each variable to be
$x_1 \in [a, b], x_2 \in [c, d]$, we can say that the image of
function $f$ will be in the interval $[a+c, b+d]$. IA defines the common mathematical operations and functions for
such intervals.
It is easy to see that IA can only give bounds for the expression
instead of an accurate range because it does not track dependencies
between arguments of operators. For instance, the IA result for the expression $x_1
- x_1$ with $x_1 \in [a, b]$ is the interval $[a - b, b -
  a]$ instead of $[0, 0]$.

IA has previously been used for SR for improving model quality as it allows to detect and
reject partially defined functions resulting e.g. from division by zero or taking the logarithm of a non-positive number.
The use of IA for SR was first proposed
in~\citep{keijzer2003improving} where IA was used to verify that a
given expression is defined within the domain spanned by the variable
values in the data set. The author showed that this indeed helps
avoiding partially defined functions even when using standard division instead of the commonly used protected division operator.

\cite{pennachin2010robust} 
used Affine Arithmetic, an extension to IA, for finding robust solutions. The
authors observed that the models generated by their algorithm were
robust w.r.t. extrapolation error.

In ~\citep{dick2017revisiting} the approach of~\citep{keijzer2003improving} 
was extended by introducing crossover and mutation operators that make use of
IA to generate feasible expressions. The
experiments showed a faster convergence rate with interval-aware operators.

We also include IA into SR. 
Our work is in fact a generalization of the idea discussed
in~\citep{keijzer2003improving} and~\citep{pennachin2010robust},
but instead of limiting the use of IA to detect
only whether a function is partial or not, we also expand it to test for
other properties such as monotonicity. Another difference is that we
assume that the considered intervals are derived from prior knowledge, instead of being inferred from the data.

\subsection{Genetic Programming for Shape-constrained Symbolic Regression}
We integrate IA into two solvers for SR. In this section we describe the
integration into tree-based GP \citep{koza1992genetic} and in the next section
we describe the integration into ITEA.

We use a tree-based GP variation with optional memetic local
optimization of SR model parameters which has produced good results for a
diverse set of regression benchmark problems \citep{KommendaGPEM2020}.

The pseudo-code for our GP algorithm for shape-constrained symbolic
regression is shown in Algorithm \ref{alg:GP}. Solution candidates are
symbolic expressions encoded as expression trees. Parameters of the
algorithm are the function set $\mathcal{F}$, the terminal set
$\mathcal{T}$, the maximum length (number of nodes) $L_{\max}$ and
maximum depth $D_{\max}$ of expression trees, the number of
generations $G_{\max}$, the population size $N$, the mutation rate
$m$, the tournament group size $p$, and the number of iterations for
memetic parameter optimization $n_{\operatorname{Opt}}$.
The algorithm uses tournament selection, generational replacement and
elitism. New solution candidates are produced via sub-tree crossover
and optional mutation. The initial population is generated with the
PTC2 algorithm~\citep{Luke2000}. All evolutionary operators respect
the length and depth limits for trees.

Shape constraints are handled during fitness evaluation. First we calculate
intervals for the function output as well as the necessary partial derivatives
using the known intervals for each of the input variables and check the output
intervals against the constraint bounds.  If a solution candidate violates any
of the shape constraints the solution candidate is assigned the worst possible
fitness value.  In a second step, the prediction error is calculated only for
the remaining feasible solution candidates. The solution candidate with the
smallest error within a group is selected as the winner in tournament
selection.

We assume that the constraints are provided via a set $C$ as given in Section \ref{shape-constrained-regression}.

\subsubsection{Fitness Evaluation}
Function \emph{\ref{alg:gp-eval}} calculates the vector of residuals from the predictions
for inputs $X$ and the observed target values $y$. We use the normalized mean of squared errors
(NMSE) as the fitness indicator. NMSE is the mean of squared residuals scaled with the
inverse variance of $y$. As described in~\citep{keijzer2003improving} we implicitly scale all symbolic regression
solutions to match the mean and variance of $y$. Therefore, we know that the worst possible
prediction  (i.e. a constant value) has an NMSE of one and we use this value for infeasible
solution candidates.

\begin{algorithm}[t!]
  \SetKwFunction{Evaluate}{Evaluate}
  \SetKwFunction{InitRandomPopulation}{InitRandomPopulation}
  \SetKwFunction{SelectTournament}{SelectTournament}
  \SetKwFunction{Crossover}{Crossover}
  \SetKwFunction{Mutate}{Mutate}
  \SetKwFunction{Optimize}{Optimize}
  \SetKwFunction{FindElite}{FindElite}
  \SetKwFunction{rand}{rand}
  \SetKwData{X}{X}
  \SetKwData{y}{$\mathbf{y}$}
  \SetKwData{C}{C}
  \SetKwData{Intervals}{Intervals}
  \SetKwArray{pop}{pop}
  \SetKwArray{err}{err}
  \SetKwArray{nextPop}{nextPop}
  \SetKwArray{nextErr}{nextErr}
  \SetKwData{parent}{parent}
  \SetKwData{child}{child}
  \SetKwData{best}{best}
  \SetKwData{expr}
  \KwData{\textbf{X} and \textbf{y} are input and target variable values, 
      \textbf{C} is the vector of constraints (operators and thresholds), 
    \Intervals are intervals for all input variables,
    \pop and \nextPop are solution candidate vectors, and 
    \err and \nextErr are error vectors of solution candidates}\\
  \pop $\leftarrow$ \InitRandomPopulation{$\mathcal{F}, \mathcal{T}, L_{\max}, D_{\max}$}\;
  \For{i $\leftarrow 1$ \KwTo N \do}{
    \err{i} $\leftarrow$ \Evaluate{\pop{i}, \X, \y, \C, \Intervals}
  }
  \For{g $\leftarrow 1$ \KwTo $G_{\max}$ \do}{
    \nextPop{1}, \nextErr{1} $\leftarrow$  \FindElite{\pop, \err}\;
    
    \For{i $\leftarrow 2$ \KwTo N \do} {
      \parent$_1 \leftarrow$ \SelectTournament{\pop, \err, $p$}\;
      \parent$_2 \leftarrow$ \SelectTournament{\pop, \err, $p$}\;
      \child $\leftarrow$ \Crossover{\parent$_1$, \parent$_2$, $L_{\max}, D_{\max}$}\;
      \If{\rand{} $< m$}{
        \Mutate{\child, $\mathcal{F}, \mathcal{T}, L_{\max}, D_{\max}$}
      }
        
      \If{$n_{\operatorname{Opt}}> 0$}{
        \Optimize{\child, $n_{\operatorname{Opt}}$, \X, \y, \C, \Intervals}
      }
        
      \nextPop{i} $\leftarrow$  \child\;
      \nextErr{i} $\leftarrow$  \Evaluate{\child, \X, \y, \C, \Intervals}
    }
    \pop $\leftarrow$  \nextPop\;
    \err $\leftarrow$  \nextErr\;
  }
  \best $\leftarrow$  \pop{1}\;
  \Return{\best}

  \caption{GP($\mathcal{F}, \mathcal{T}, L_{\max}, D_{\max}, G_{\max}, N, m, p, n_{\operatorname{Opt}}$)}
  \label{alg:GP}
\end{algorithm}

\subsubsection{Optional Local Optimization}\label{sec:local-opt}
Procedure \emph{\ref{alg:gp-optimize}} locally improves the vector of numerical coefficients $\theta \in \mathbb{R}^{\text{dim}}$ of
each model using non-linear least-squares fitting using the Levenberg-Marquardt algorithm.
It has been demonstrated that gradient-based local
improvement improves symbolic regression performance \citep{Topchy2001,
  KommendaGPEM2020}. Here we want to investigate whether it can also be used in
combination with shape constraints. To test the algorithms with and without local
optimization we have included the number
of local improvement iterations $n_\text{Opt}$ as a parameter for the GP
algorithm.

\begin{function}[t!]
  \SetKwData{nmse}{nmse}
   $V \leftarrow \{ c \in C \mid \text{evalConstraint}(c, \text{Intervals}) > 0\} $\;
   \eIf(\tcc*[f]{return worst nmse for expr with violations}){V $\neq \varnothing$}{ 
     \nmse $\leftarrow 1$
   }{
     s $\leftarrow \|y - \operatorname{mean}(y)\|^2$ \tcc*[r]{normalization factor}
     \nmse $\leftarrow \min( s^{-1} \|\text{eval}(\text{expr}, X) - \mathbf{y}\|^2, 1)$ \tcc*[r]{worst nmse is 1}
   }
   \Return{\nmse}
\caption{Evaluate(expr, $X, \mathbf{y}$, C, Intervals)}
\label{alg:gp-eval}
\end{function}

The local improvement operator extracts the initial parameter values $\theta_{\text{init}}$
from the solution candidates and updates the values after
optimization. Therefore, improved parameter values $\theta^\star$ become part of the
genome and can be inherited to new solution candidates (Lamarckian
learning~\citep{Houck1997}).

\begin{procedure}
  \SetKwData{expr}{expr}
  \SetKwFunction{ExtractParam}{ExtractParam}
  \SetKwFunction{UpdateParam}{UpdateParam}
  $\mathbf{\theta}_{\text{init}} \leftarrow$ \ExtractParam{\expr} \tcc*[r]{use initial params from expr}
    fx $\leftarrow (\mathbf{\theta}_i) \Rightarrow$
    \Begin( \tcc*[f]{eval expression for $\theta_i$}){\UpdateParam{\expr,$\mathbf{\theta}_i$}\;
        $\operatorname{eval}(\expr, X)$
      }
      $\mathbf{\theta}^\star \leftarrow \operatorname{LevenbergMarquardt}(\mathbf{\theta}_{\text{init}}, n_{\operatorname{Opt}}, \mathbf{y}, \text{fx})$\;      
  \UpdateParam{\expr, $\theta^\star$}\;
  \caption{Optimize(expr, $n_{\operatorname{Opt}}, X, \mathbf{y}$ C, Intervals)}
  \label{alg:gp-optimize}
\end{procedure}

\subsection{Interaction-Transformation Evolutionary
Algorithm}\label{interaction-transformation-evolutionary-algorithm}
The second SR method that we have extended for shape-constrained SR is
the Interaction-Transformation Evolutionary Algorithm (ITEA). ITEA is
an evolutionary algorithm that relies on mutation to search for an
optimal Interaction-Transformation expression~\citep{de2018greedy}
that fits a provided data set. The Interaction-Transformation (IT)
representation restricts the search space to simple expressions
following a common pattern that greatly simplifies optimization
of model parameters and handling shape constraints.
The IT representation specifies a pattern of function forms:

\begin{equation*}
\hat{f}(x) = \sum_{i}{w_i \cdot t_i(\prod_{j=1}^{d}{x_j^{k_{ij}}})},
\end{equation*}

\noindent where $w_i$ is the weight of the $i$-th term of the linear
combination, $t_i$ is called a transformation function and can be any 
univariate function that is reasonable for the studied system
and $k_{ij}$ is the strength of interaction for the $j$-th variable
on the $i$-th term.

The algorithm is a simple mutation-based evolutionary algorithm depicted
in Algorithm ~\ref{alg:itea}. It starts with a random population of expressions
and repeats mutation and selection until convergence.

\begin{algorithm}[tb!]
  \SetKwData{P}{pop}\SetKwData{F}{children}\SetKwData{f}{child}
  \SetKwFunction{GenRandomPopulation}{GenRandomPopulation}\SetKwFunction{Mutate}{Mutate}\SetKwFunction{Select}{Select}
  \SetKwInOut{Input}{input}\SetKwInOut{Output}{output}
  \Input{data points $X$ and corresponding set of target variable $\mathbf{y}$.}
\Output{symbolic function $f$}
\BlankLine

\P $\leftarrow$ \GenRandomPopulation{}\;

\While{stopping criteria not met}{
	\F $\leftarrow$ \Mutate{\P}\;
        \P $\leftarrow$ \Select{\P, \F}\;
}
\Return $\operatorname{arg} \max_{p.fit} $ \P\;
	\caption{ITEA~\citep{de2019interaction}.}
\label{alg:itea}
\end{algorithm}

The mutation procedure chooses one of the following actions at random:
\begin{itemize}
    \item Remove one random term of the expression.
    \item Add a new random term to the expression.
    \item Replace the interaction strengths of a random term of the expression.
    \item Replace a random term of the expression with the positive interaction of another random term.
    \item Replace a random term of the expression with the negative interaction of another random term.
\end{itemize}

The positive and negative interactions of terms is the sum or subtraction
of the strengths of the variables. For example, if we have two terms
$x_1^2x_2^3$ and $x_1^{-1}x_2^1$, and we want to apply the positive interaction
to them, we would perform the operation $x_1^2 \cdot x_1^{-1} \cdot x_2^3 \cdot x_2^1$
that results in $x_1x_2^4$, similarly with the negative interactions we divide the first
polynomial by the other, so performing the operation $x_1^2 \cdot x_1^1 \cdot x_2^3 \cdot x_2^{-1}$,
resulting in $x_1^3x_2^2$.

Weights within IT expressions can be efficiently optimized using ordinary
least squares (OLS) after the mutation step.

This algorithm was reported to surpass different variants of GP
 and stay on par with nonlinear regression algorithms~\citep{de2019interaction}.

\subsection{Feasible-Infeasible Two-population with
Interaction-Transformation}\label{feasible-infeasible-two-population-with-interaction-transformation}

In order introduce the shape-constraints into ITEA, 
we will use the approach called Feasible-Infeasible Two-population
(FI-2POP)~\citep{kimbrough2008feasible} that works with two populations: 
one for feasible models and another for infeasible models. Despite its simplicity,
it has been reported to work well on different
problems~\citep{liapis2015procedural,scirea2016metacompose,covoes2018classification}. 
We have chosen this approach to deal with constraints for the ITEA algorithm since it
does not demand fine tuning of a penalty function and a penalty coefficient, does not require
any change to the original algorithm (as the repairing and constraint-aware operators),
and does not introduce a high computational cost (i.e., multi-objective optimization).

Specifically for ITEA, the adaptation (named FI-2POP-IT) is depicted in
Algorithm~\ref{alg:fi2pop}.  
The main difference from Algorithm~\ref{alg:itea} is that in FI-2POP-IT every function
now produces two distinct populations, one for feasible solutions and another for
the infeasible solutions. For example, the mutation operator when applied to an individual
can produce either a feasible solution or an infeasible one, this new individual will be
assigned to its corresponding population.

The fitness evaluation differs for each population, the feasible population minimizes 
the error function while the infeasible population minimizes the constraint violations.

\begin{algorithm}[tb!]
	\SetKwData{F}{feas}\SetKwData{I}{infeas}\SetKwData{Fp}{child-feas}\SetKwData{Ip}{child-infeas}
	\SetKwData{f}{f}\SetKwData{i}{i}
  \SetKwFunction{GenRandom}{GenRandomPopulation}\SetKwFunction{Mutate}{Mutate}\SetKwFunction{Select}{Select}
  \SetKwInOut{Input}{input}\SetKwInOut{Output}{output}
  \Input{data points $X$ and corresponding set of target variable $\mathbf{y}$.}
  \KwData{\F is the feasible population, \I is the infeasible population, \Fp and \Ip are the feasible and infeasible child population, respectively.}
\Output{symbolic function $f$}
\BlankLine

\F, \I $\leftarrow$ \GenRandom{}\;

\While{stopping criteria not met}{
    \Fp, \Ip $\leftarrow$ \Mutate{\F $\cup$ \I}\;
    \F, \I $\leftarrow$ \Select{\F $\cup$ \Fp $\cup$ \I $\cup$ \Ip}\;
}
\Return $\operatorname{arg} \max_{s \in \F} $ fitness($s$)\;
	\caption{FI-2POP-IT: Interaction-Transformation Evolutionary Algorithm with Feasible-Infeasible Two-Population.}
\label{alg:fi2pop}
\end{algorithm}

\subsubsection{Defining Constraints with Interval
Arithmetic}\label{defining-constraints-with-interval-arithmetic}

The IT representation allows for a simple algorithm to calculate the
$n$-th order partial derivatives of the expression and to
perform IA. Additionally, only linear real-valued parameters are
allowed for IT expressions. Therefore
parameters can be optimized efficiently using ordinary least squares
(OLS) and all shape constraints can be transformed to functions which
are linear in the parameters.

Given an IT expression and the set of domains for each of the input
variables, it is possible to calculate the image of the expression as the
sum of the images of each term multiplied by their corresponding weight.
The image of an
interaction term $Im(.)$ is the product of the exponential of each variable interval $D(.)$ 
with the corresponding strength:

\begin{equation*}
Im(p(x)) = \prod_{i=1}^{d}{D(x_i)^{k_i}},
\end{equation*}

\noindent 
Arithmetic operators are well defined in
interval arithmetic and the  images of univariate non-linear functions can
be easily determined.

The derivative of an IT expression can be calculated by the chain rule
as follows ($g_i = t_i \circ p_i$):

\begin{align*}
\frac{\partial IT(x)}{\partial x_j} &= w_1 \cdot g'_1(x) + \ldots + w_n \cdot g'_n(x) \\
\frac{\partial g_i(x)}{\partial x_j} &= t'_i(p_i(x)) \cdot  k_j\frac{p_i(x)}{x_j} 
\end{align*}

All of the aforementioned transformation functions have derivatives
representable by the provided functions. Following the same rationale the second order 
derivatives can be calculated as well.

\section{Experimental Setup}\label{experimental-setup}

For testing the effectiveness of the proposed approach we have
created some artificial data sets generated from fluid dynamics
engineering (FDE) taken from~\citep{chen2018multilevel} and some selected physics models from the \emph{Feynman Symbolic Regression Database}\footnote{\url{https://space.mit.edu/home/tegmark/aifeynman.html}}~\citep{udrescu2020ai} (FEY). Additionally, we have used 
 data sets built upon real-world measurements from physical systems 
 and engineering (RW).
We have chosen these data sets because they are
prototypical examples of non-trivial non-linear models which are
relevant in practical applications.

For each of the problem instances we have defined a set of shape constraints
that must be fulfilled by the SR solutions. For the FDE and FEY data sets,
we have used constraints that can be derived from the known expressions over the given input space. For the physical systems we have defined the shape constraints based on expert knowledge. The input space and monotonicity constraints for all problem instances are given in the supplementary material.

In the following subsections we give more details on the problem
instances including the constraint definitions and describe the
methods to test the modified algorithms.

\subsection{Problem Instances}\label{problem-instances}

All the data sets are described by the input variables names, the target variable,
the true expression if it is known, domains of the variables for the
training and test data, expected image of the function and
monotonic constraints. All data set specifications are provided as a supplementary material.

Details for \emph{Aircraft Lift}, \emph{Flow Psi}, \emph{Fuel Flow} models can be found in ~\citep{anderson2010fundamentals, anderson1982modern}. 
Additionally, we have selected a subset of the FEY
problem instances \citep{udrescu2020ai}. The selection criteria was
those problem instances which have been reported as unsolved 
without dimensional analysis or for which a required
runtime of more than five minutes was reported by \cite{udrescu2020ai}.
We have not included the three benchmark problems from the closely related work of \cite{Bladek}, because we could easily solve them in preliminary experiments and additionally, the results would not be directly comparable because we do not support symmetry constraints.

All constraints for FDE and FYE have been
determined
by analysing the known expression and its partial derivatives.
The formulas for the FDE and FYE problem instances are shown in Table \ref{tab:problem-instances}.
For all these problem instances we have generated
data sets from the known formula by randomly sampling the input space 
using $100$ data points for training and $100$ data
points for testing. For each of the 
data sets we generated two versions: one without noise and one where we have added normally
distributed noise to the target variable $y' = y + N(0, 0.05
\sigma_y)$. Accordingly, the optimally achievable NMSE
for the noisy problem instances is $0.25\%$.

We have included three real-world data sets (RW). 
The dataset for \emph{Friction $\mu_{\operatorname{stat}}$} and \emph{Friction $\mu_{\operatorname{dyn}}$} has been collected in a
standardized testing procedure for friction plates and has two target variables that we model independently. The \emph{Flow Stress} data set
stems from hot compression tests of aluminium cylinders \citep{Kabliman2019}.
The constraints for these data sets where set by specialists in the field. 

 The \emph{Cars} data set has values for displacement, horsepower, weight, acceleration time and fuel efficiency of almost 400 different car models.
 This set was first used by~\cite{mammen2001general} and then evaluated
 by~\cite{shah2016soft} to assess the performance of Support Vector Regression with soft constraints.

\begin{table}[t]
  \begin{tabular}{p{5cm}>{$}c<{$}}
    Name & \text{Formula} \\
    \hline
    Aircraft lift & C_L = C_{L\alpha}(\alpha + \alpha_0) + C_{L\delta_e}\delta_e\frac{S_{\operatorname{HT}}}{S_{\operatorname{ref}}} \\
    Flow psi & \Psi = V_\infty r \sin (\frac{\theta}{2\pi})\left(1 - \left(\frac{R}{r}\right)^2 \right) + \frac{\Gamma}{2\pi}\log\frac{r}{R} \\
    Fuel flow &   \dot m = \frac{p_0 A\star}{\sqrt{T_0}}\sqrt{\frac{\gamma}{R}\left(\frac{2}{1+\gamma} \right)^{(\gamma + 1)/(\gamma - 1)}} \\

    Jackson 2.11 & \frac{q}{4 \pi \epsilon {{y}^{2}}}\, \left( 4 \pi \epsilon \mathit{Volt}\, d-\frac{q d\, {{y}^{3}}}{{{\left( {{y}^{2}}-{{d}^{2}}\right) }^{2}}}\right) \\
    Wave power & \frac{-32}{5}\frac{G^4}{c^5} \frac{{{\left( \mathit{m1}\, \mathit{m2}\right) }^{2}}\, \left( \mathit{m1}+\mathit{m2}\right)}{r^5} \\
    I.6.20 & \operatorname{exp}\left( \frac{-{{\left( \frac{\theta}{\sigma}\right) }^{2}}}{2}\right) \frac{1}{\sqrt{2 \pi} \sigma} \\
    I.9.18 & \frac{G\, \mathit{m1}\, \mathit{m2}}{{{\left( \mathit{x2}-\mathit{x1}\right) }^{2}}+{{\left( \mathit{y2}-\mathit{y1}\right) }^{2}}+{{\left( \mathit{z2}-\mathit{z1}\right) }^{2}}} \\
    I.15.3x & \frac{x-u t}{\sqrt{1-\frac{{{u}^{2}}}{{{c}^{2}}}}}  \\
    I.15.3t & \left(t-\frac{u x}{{{c}^{2}}}\right)\frac{1}{\sqrt{1-\frac{{{u}^{2}}}{{{c}^{2}}}}} \\
    I.30.5 & \operatorname{asin}\left( \frac{\mathit{lambd}}{n d}\right) \\
    I.32.17 & \frac{1}{2} \epsilon c\, {{\mathit{Ef}}^{2}}\, \frac{8 \pi {{r}^{2}}}{3}\, \frac{{{\omega}^{4}}}{{{\left( {{\omega}^{2}}-{{{{\omega}_0}}^{2}}\right) }^{2}}} \\
    I.41.16 & \frac{h\, {{\omega}^{3}}}{{{\pi}^{2}}\, {{c}^{2}}\, \left( \operatorname{exp}\left( \frac{h \omega}{\mathit{kb} T}\right) -1\right) } \\
    I.48.20 & \frac{m\, {{c}^{2}}}{\sqrt{1-\frac{{{v}^{2}}}{{{c}^{2}}}}} \\
    II.6.15a & \frac{\frac{{p_d}}{4 \pi \epsilon} 3 z}{{{r}^{5}}} \sqrt{{{x}^{2}}+{{y}^{2}}} \\
    II.11.27 & \frac{n \alpha}{1-\frac{n \alpha}{3}} \epsilon \mathit{Ef} \\
    II.11.28 & 1+\frac{n \alpha}{1-\frac{n \alpha}{3}} \\
    II.35.21 & {n_{\mathit{rho}}} \mathit{mom} \operatorname{tanh}\left( \frac{\mathit{mom} B}{\mathit{kb} T}\right)  \\
    III.9.52 & \frac{{p_d} \mathit{Ef} t}{h} {{\sin{\left( \frac{\left( \omega-{{\omega}_0}\right)  t}{2}\right) }}^{2}} \frac{1}{{{\left( \frac{\left( \omega-{{\omega}_0}\right)  t}{2}\right) }^{2}}}  \\
    III.10.19 & \mathit{mom}\, \sqrt{{{\mathit{Bx}}^{2}}+{{\mathit{By}}^{2}}+{{\mathit{Bz}}^{2}}} \\
  \end{tabular}
  \caption{\label{tab:problem-instances}Synthetic problem instances used for testing. The first three functions are from the FDE data sets, the rest from the FEY data sets.}
\end{table}

\subsection{Testing Methodology}

Each one of the following
experiments has been repeated $30$ times and, for every repetition, we have
measured the NMSE in percent for both training and test data.
The algorithms receive only the training
set as the input and only the best found model is applied to the test set.
Runs with and without shape constraints have been executed on the same
hardware.

The first experiment tests the performance of the unmodified algorithms, i.e.,
without shape constraints.  This test verifies whether unmodified SR algorithms
identify conforming models solely from the observed data. If this result turns
out negative, it means that constraint handling mechanisms are required for
finding conforming models.

In the second experiment, we test the performance of the two modified
algorithms. This will verify whether the algorithms are able to identify models
conforming to prior knowledge encoded as shape constraints.

This testing approach allows us to compare the prediction errors (NMSE) of
models produced with and without shape constraints to assess whether the
prediction error improves or deteriorates.
We perform the same analysis for the two groups of instances with and without noise separately. 

\subsection{Algorithms Configurations}

 In the following, the abbreviation GP refers to tree-based GP without
 local optimization and GPC refers to tree-based GP with local
 optimization. Both algorithms can be used with and without shape
 constraints. Table \ref{tab:algorithm-configuration} shows
 the parameter values that have been used for the experiments with GP
 and GPC.  ITEA refers to the Interaction-Transformation Evolutionary
 Algorithm and FI-2POP-IT (short form: FIIT) refers to the two-population version of ITEA
 which supports shape constraints. Table~\ref{tab:itea-configuration}
 shows the parameters values for ITEA and FI-2POP-IT.

\begin{table}[t!]
  \centering
  \begin{tabular}{lp{8cm}}
    Parameter & Value \\
    \toprule
    Population size & 1000 \\
    Generations  & $200$ \\
     & $20$ (for GPC with memetic optimization)\\
    Initialization & PTC2 \\
    $L_{max}$, $D_{max}$ & 50, 20 \\    
    Fitness evaluation & NMSE with linear scaling \\
    Selection & Tournament (group size = 5)\\
    Crossover & Subtree crossover \\
    Mutation (one of) & Replace subtree with random branch \newline Add $x\sim N(0, 1)$ to all numeric parameters \newline Add $x\sim N(0, 1)$ to a single numeric parameter\newline Change a single function symbol  \\ 
    Crossover rate & 100\% \\
    Mutation rate & 15\%\\
    Replacement & Generational with a single elite\\
    Terminal set & real-valued parameters and input variables\\
    Function set & $+, *, \%, \log, \exp, \sin, \cos, \tanh, x^2, \sqrt{x} $\\
    GPC & max. 10 iterations of Levenberg-Marquardt (LM)\\    
    \bottomrule
  \end{tabular}
  \caption{\label{tab:algorithm-configuration}GP parameter
    configuration used for the experiments (\% refers to protected
    division). For the GPC runs we need fewer generations because of the faster convergence with local optimization.}
\end{table}

\begin{table}[t!]
  \centering
  \begin{tabular}{lp{8cm}}
  Parameter & Value \\
    \toprule
    Population size & $200$ \\
    Number of Iterations & $500$ \\
    Function set & $\sin, \cos, \tanh, \sqrt, \log, \log1p, \exp$ \\
    Fitness evaluation & RMSE \\    
    Maximum number of terms (init. pop.) & $4$ \\
    Range of strength values (init. pop.) & $[-4, 4]$ \\
    Min. and max. term length & $2, 6$ \\
    Regression model & OLS \\  
    \bottomrule
  \end{tabular}
  \caption{\label{tab:itea-configuration}Parameter values for the
    ITEA algorithm that have been used for all experiments.}
\end{table}

For comparison we have used auto-sklearn \citep{FeurerAutoSklearn} (AML) and
an implementation of shape-constrained polynomial regression (SCPR) \citep{curmei2020shapeconstrained}\footnote{We have initially
also used XGboost, a popular implementation of gradient boosted trees,  because it has previously been found that it compares well with symbolic regression methods
\citep{PennML:WhereAreWeNow} and it supports monotonicity constraints. However, the results were much worse compared to the symbolic regression results which 
indicates that the method is not particularly well-suited for the problem instances in the benchmark set which are smooth and of low dimensionality.}.
Auto-sklearn does not allow to set monotonicity constraints. Therefore, we can only use
it as a comparison for symbolic regression models without constraints.
We execute 30 independent repetitions for each problem with
each run limited to one hour.

The degree for SCPR has been determined for each problem instance
using 10-fold cross-validation and using the best CV-RMSE as the
selection criteria. We tested homogeneous polynomials up to degree eight
(except for I.9.18 where we used a maximum degree of five because of
runtime limits). The runtime for the grid search was limited to one
hour for each problem instance. The final model is then trained with
the selected degree on the full training set. SCPR is a deterministic
algorithm therefore only a single run was required for each problem
instance.

\section{Results}\label{results}

In this section we report the obtained results in tabular 
and graphical forms focusing on the violations of the shape constraints,
goodness of fit, computational overhead, and
extrapolation capabilities with and without shape constraints.

\subsection{Constraint Violations}\label{const_viol}
The procedure for checking of constraint violations is as
follows: for each problem instance we sample a million points uniformly from the full input
space to produce a data set for checking constraint violations.
The final SR solutions as well as their partial
derivatives are then evaluated on this data set. If there is at least
one point for which the shape constraints are violated then we count
the SR solution as infeasible.

\begin{figure}[t!]
\centering
\begin{subfigure}[b]{0.3\textwidth}
\includegraphics[width=\textwidth]{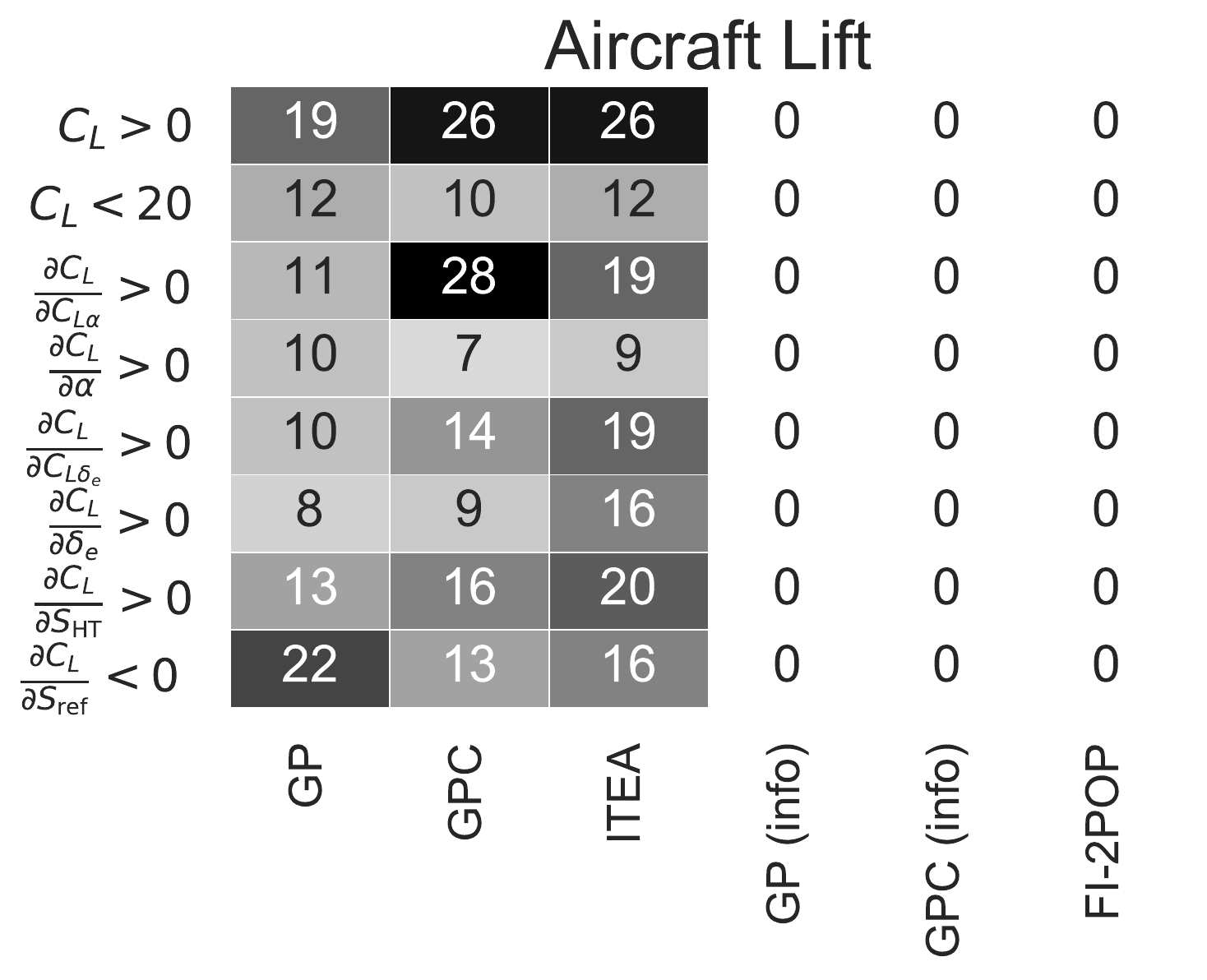}
\caption{}
\label{fig:aircraft_violations}
\end{subfigure}
\begin{subfigure}[b]{0.3\textwidth}
\includegraphics[width=\textwidth]{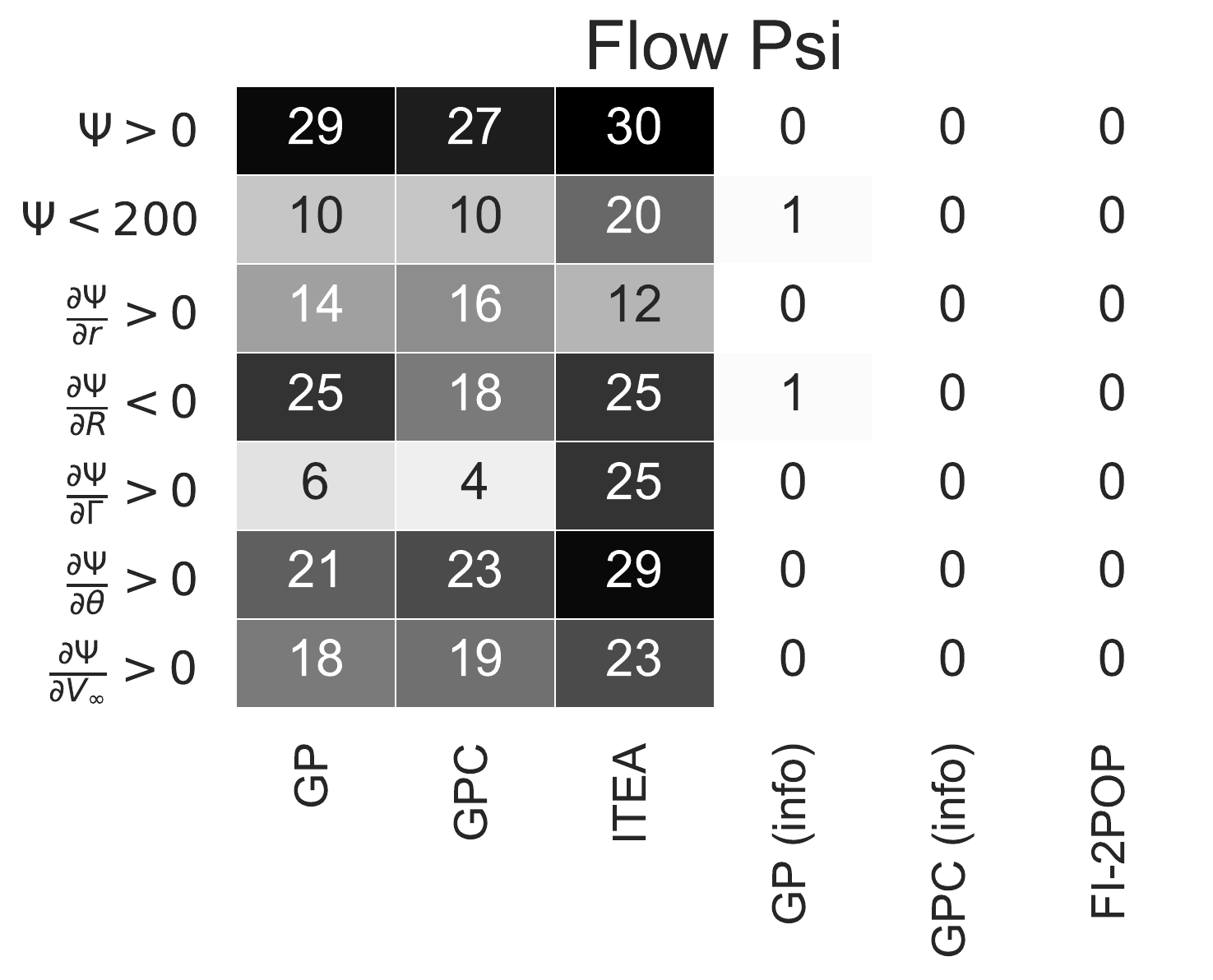}
\caption{}
\label{fig:flowpsi_violations}
\end{subfigure}
\begin{subfigure}[b]{0.3\textwidth}
\includegraphics[width=\textwidth]{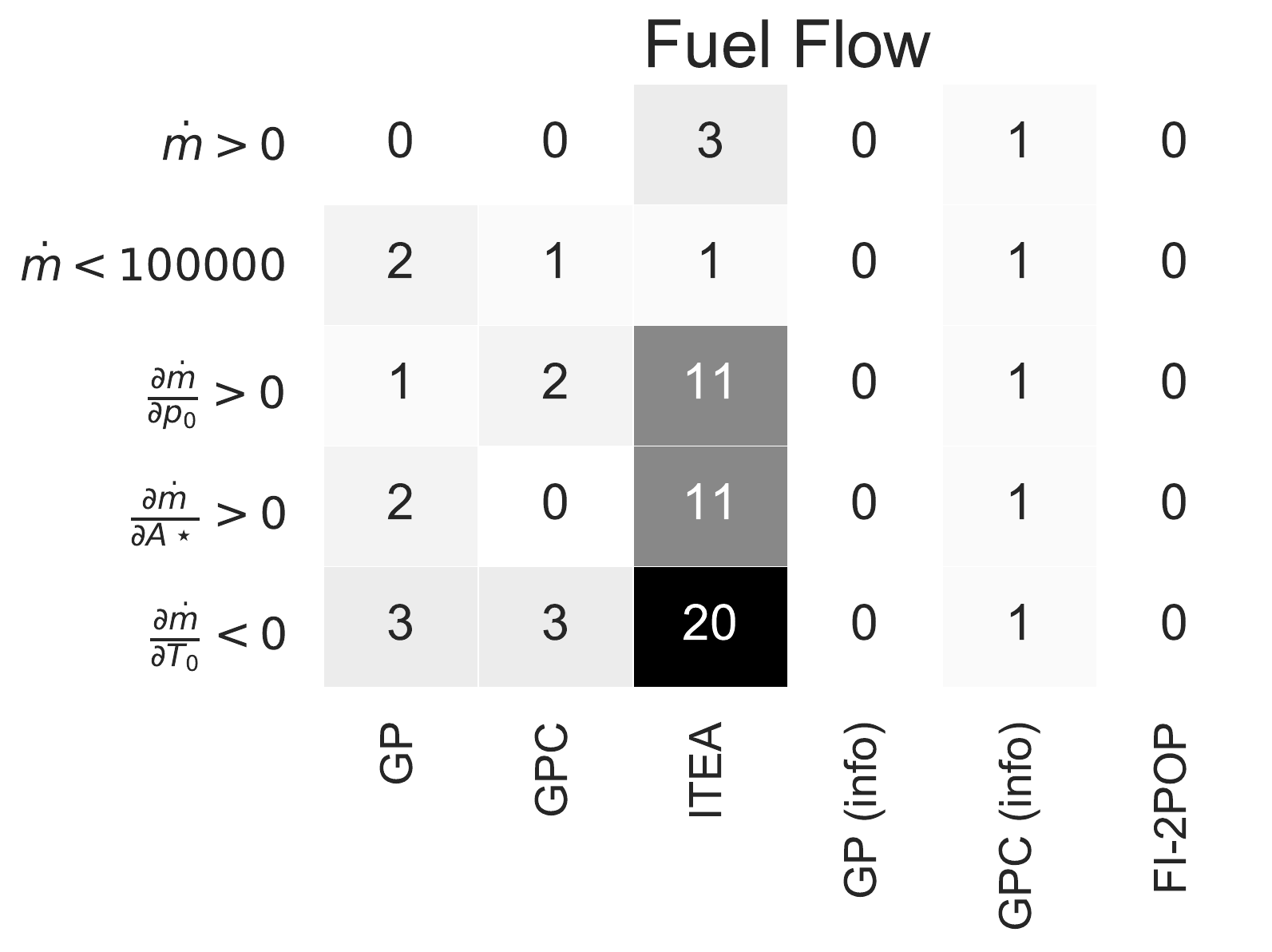}
\caption{}
\label{fig:fuelflow_violations}
\end{subfigure}
\vspace{0.5em}

\label{fig:const_viols}
\caption{Constraint violation frequency for the solutions of every
  algorithm over the $30$ executions for the FDE data sets.
  Only for the \emph{Fuel Flow} problem a  feasible solution can be
  identified even without shape constraints. 
  FI-2POP always produces feasible solutions due to the nature of its constraint
  handling mechanism.}
\end{figure}

Figures~\ref{fig:aircraft_violations}~to~\ref{fig:fuelflow_violations} show
exemplarily the number of infeasible SR models for each algorithm and
constraint, over the $30$ runs for the FDE data sets.
The picture is similar for all problem instances. We observe that without shape constraints none of the algorithms produce feasible solutions in every run. 

Only for the easiest data sets (e.g. \emph{Fuel Flow}) we re-discovered the data-generating functions -- which
necessarily conform to the constraints -- in multiple runs. 
Over all data sets, ITEA has the highest probability of producing
infeasible models. We observe that even a small amount of noise
increases the likelihood to produce infeasible models.  In short, we can see
that we cannot hope to find feasible solutions using SR algorithms without
shape constraints for noisy data sets. This is consistent with the observation in~\citep{Bladek}.

The results for the extended algorithms show, that with few exceptions, all three SR algorithms
are capable of finding feasible solutions most of the time.

\subsection{Interpolation and Extrapolation}
The difference between SR models and shape-constrained SR models becomes
clearer when we visualize their outputs.

\begin{figure}  
   \begin{subfigure}[b]{0.5\textwidth}
   \includegraphics[width=\textwidth]{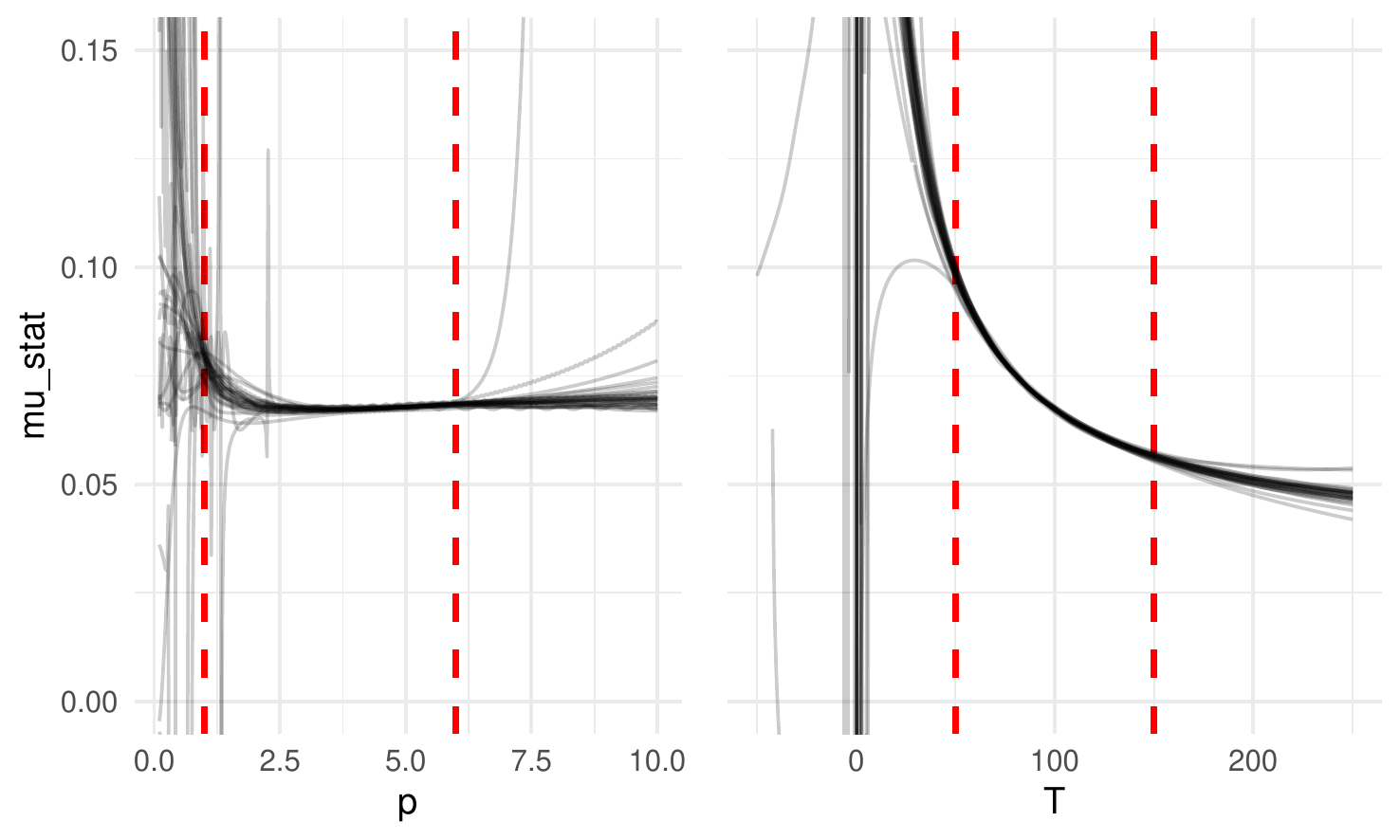}
   \caption{GP}
   \label{fig:gpe-noinfo}
   \end{subfigure}
   \begin{subfigure}[b]{0.5\textwidth}
   \includegraphics[width=\textwidth]{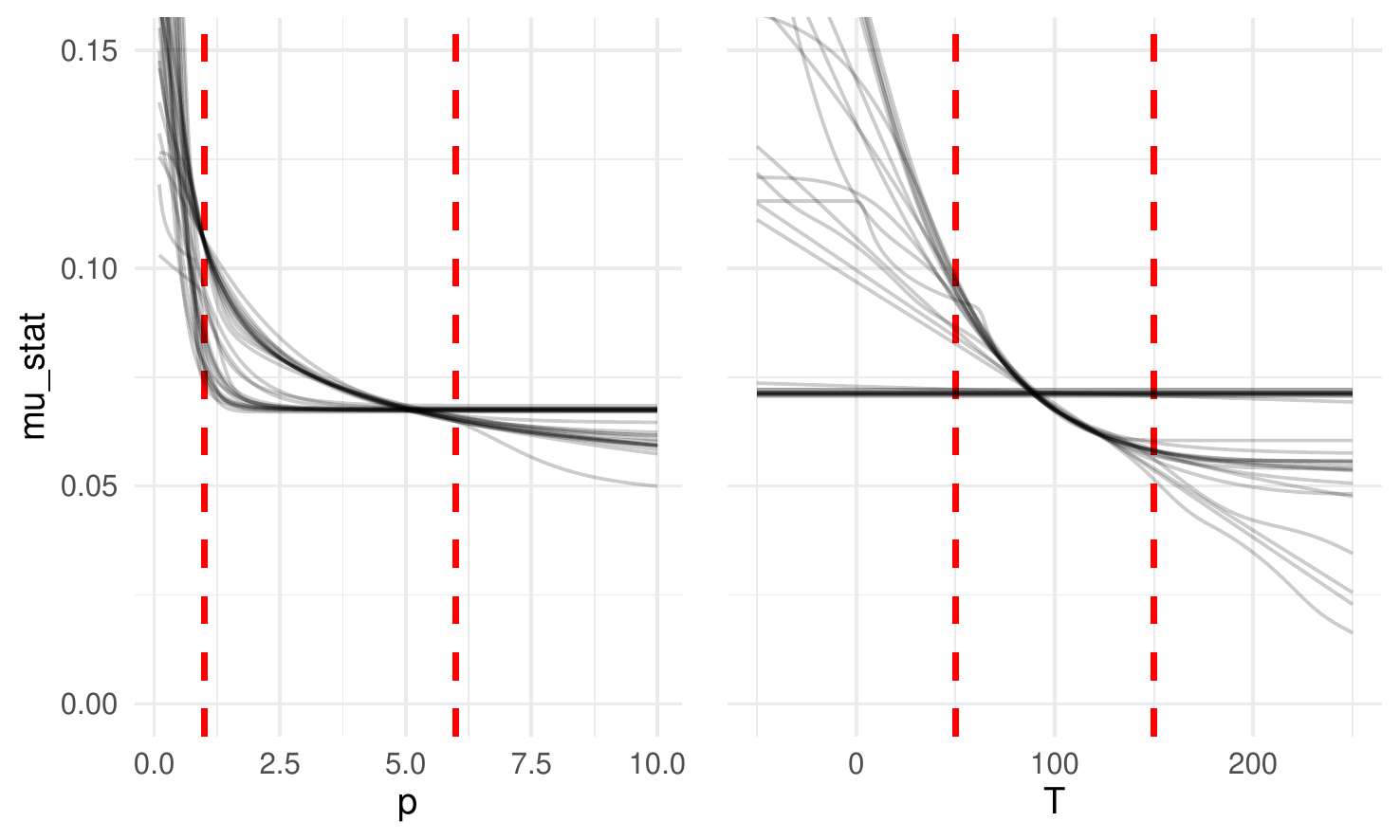}
   \caption{GP (info)}
   \label{fig:gpe-info}
   \end{subfigure}
   \vspace{0.3em}
   
   \begin{subfigure}[b]{0.5\textwidth}
   \includegraphics[width=\textwidth]{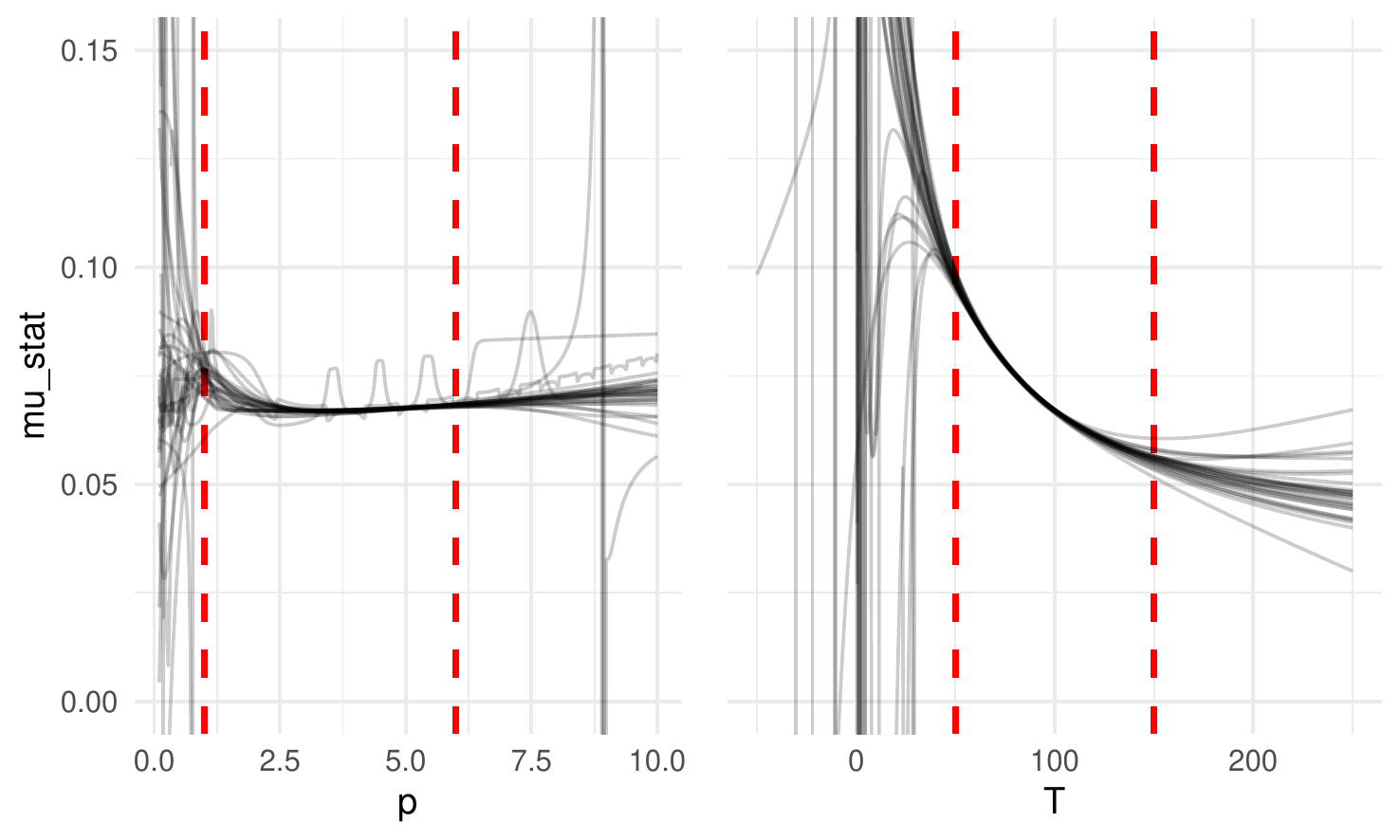}
   \caption{GPC}
   \label{fig:gpce-noinfo}
   \end{subfigure}
   \begin{subfigure}[b]{0.5\textwidth}
   \includegraphics[width=\textwidth]{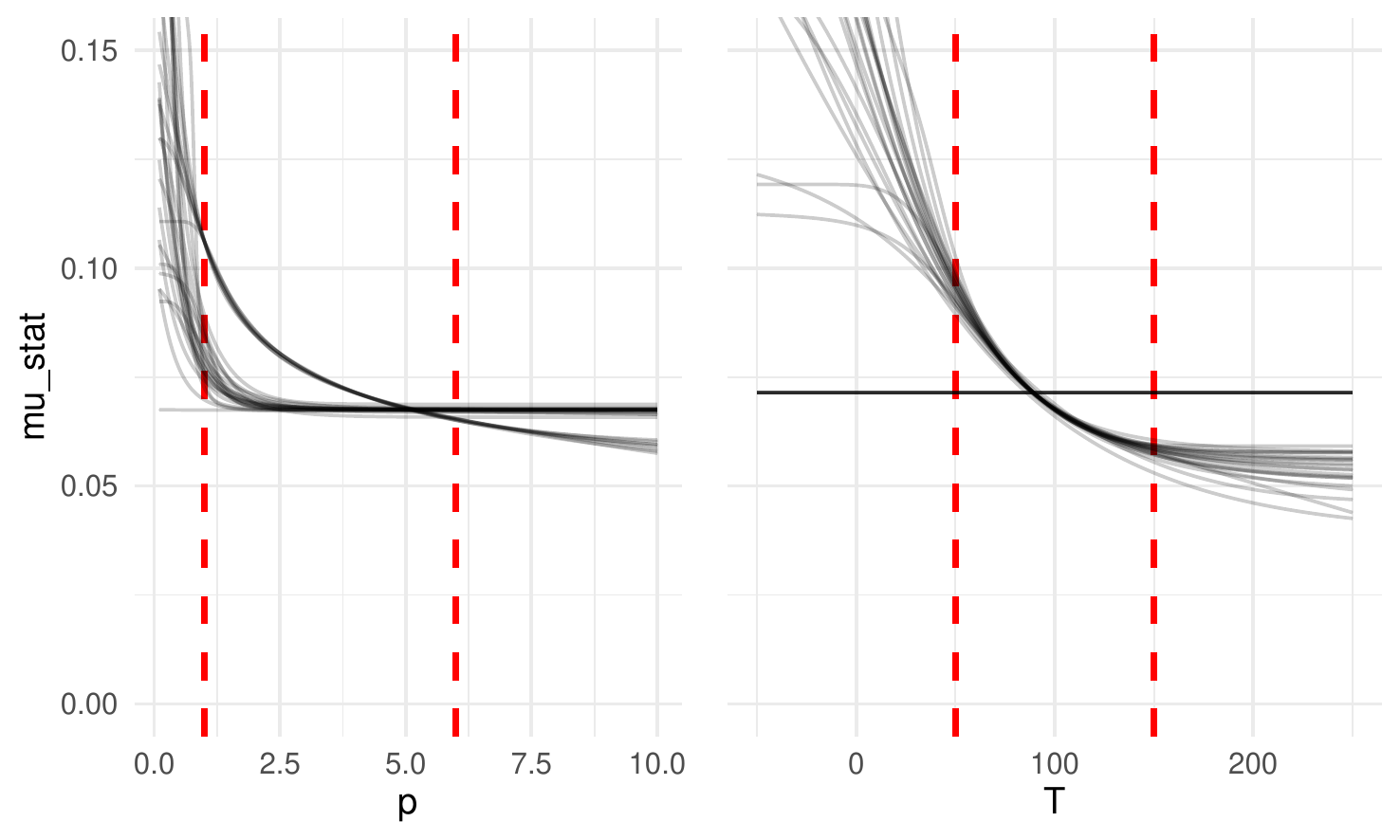}  
   \caption{GPC (info)}
   \label{fig:gpce-info}
   \end{subfigure}  
   \vspace{0.3em}
   
   \begin{subfigure}[b]{0.5\textwidth}
   \includegraphics[width=\textwidth]{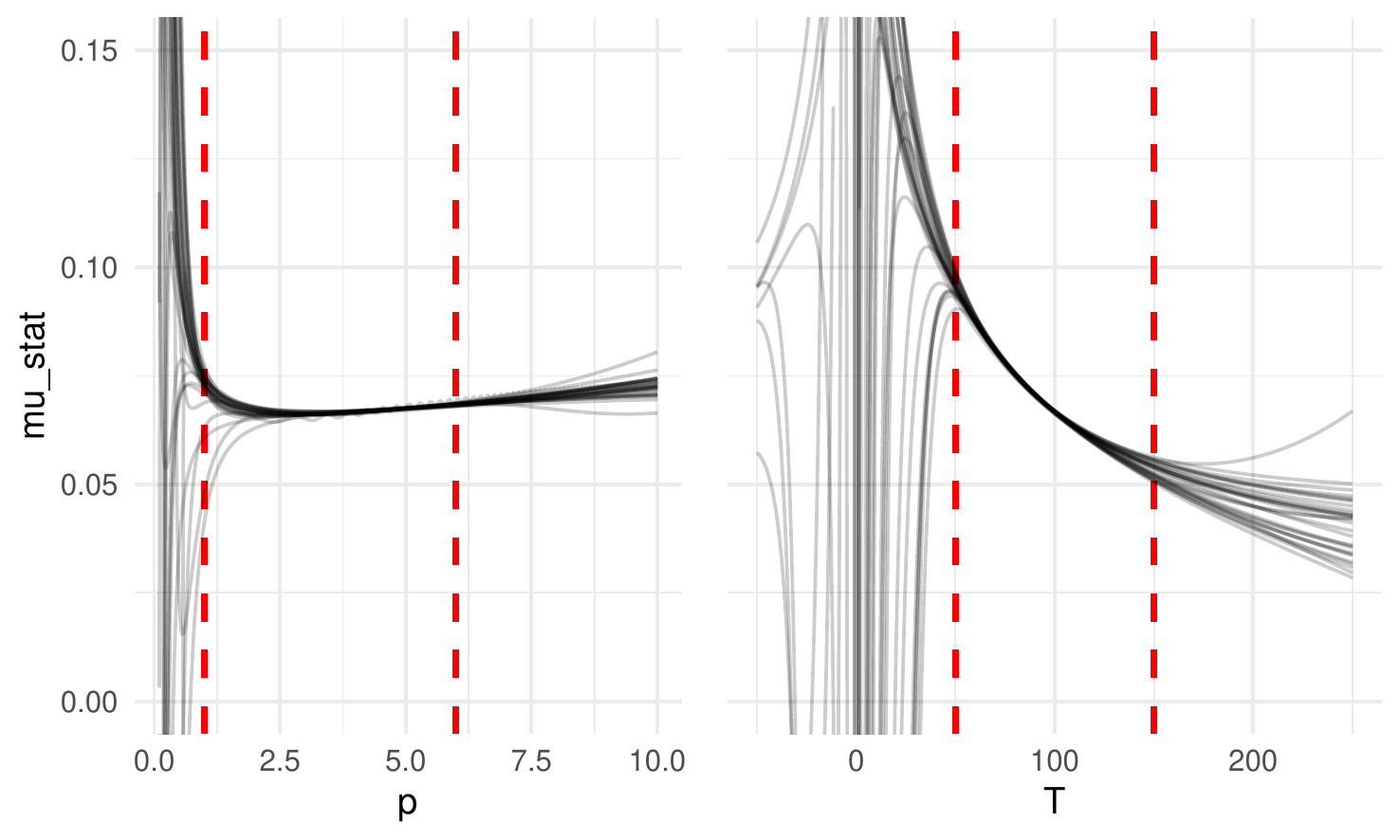} 
   \caption{ITEA}
   \label{fig:iteae-noinfo}
   \end{subfigure}  
   \begin{subfigure}[b]{0.5\textwidth}
   \includegraphics[width=\textwidth]{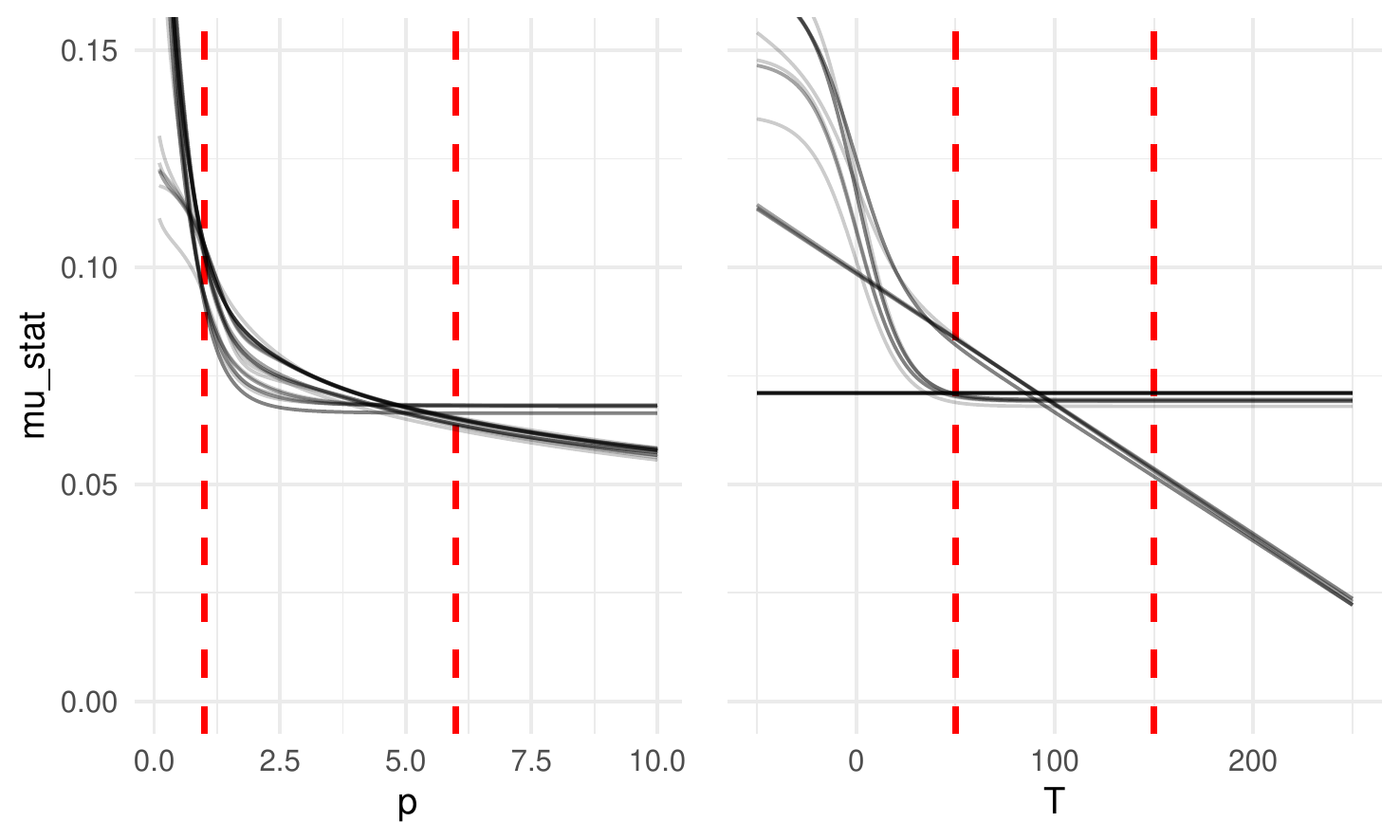}
   \caption{FI-2POP-IT}
   \label{fig:iteae-info}
   \end{subfigure}
 
 \label{fig:flowpsi-extrapolation}
 \caption{Partial dependence plots for the Friction
   $\mu_{\operatorname{stat}}$ models found by each algorithm over the
   $30$ runs. Dashed lines mark the subspace from which training and
   test points were sampled. Algorithms with shape constraints (b, d,
   f) produce SR solutions which conform to prior knowledge and have
   better extrapolation behaviour but increased prediction error (cf. Table \ref{tab:nmse-test}).}
 \end{figure}

Figures~\ref{fig:gpe-noinfo}~to~\ref{fig:iteae-info} show the predictions of
every SR model identified for the  \emph{Friction} $\mu_\text{stat}$  data
set. The partial dependence plots show the predictions for $\mu_\text{stat}$
for all 30 models over the complete range of values allowed for the two
variables $p$ and $T$. The plot of $\mu_\text{stat}$ over $p$ shows the
predictions when $v$ and $T$ are fixed to their median values and the plot of $\mu_\text{stat}$ over
$T$ shows the predictions when $p$ and $v$ are fixed. The dashed vertical lines
mark the subspace from which we have sampled points for the training and test
data sets.

The algorithms without shape constraints produce extreme predictions
when extrapolating (Figures \ref{fig:gpe-noinfo}, \ref{fig:gpce-noinfo}, \ref{fig:iteae-noinfo}). For instance many of the functions have poles at a
temperature $T$ close to zero which are visible in the plots as
vertical lines. GP and GPC without shape constraints produced a few
solutions which are wildly fluctuating over $p$ even within the
interpolation range. Within the interpolation range ITEA produced the
best SR solutions for the \emph{Friction} data sets (see Figure~\ref{fig:iteae-noinfo}
as well as Table~\ref{tab:nmse-test}). However, the models do not
conform to prior knowledge as we would expect that $\mu_\text{stat}$
decreases with increasing pressure and temperature and the models show a slight increase in $\mu_{\operatorname{stat}}$ when increasing $p$.

The model predictions for shape-constrained SR are shown in
Figures~\ref{fig:gpe-info},~\ref{fig:gpce-info}, and
\ref{fig:iteae-info}. The visualization clearly shows that there is
 higher variance and that all algorithms produced a few bad
or even constant solutions. This invalidates our hypothesis that shape-constrained SR leads to improved predictive accuracy and instead indicates that there
are convergence issues with our approach of including shape constraints. We will discuss this in more detail in the following sections. The visualization also shows that the solutions with shape constraints
have better extrapolation behaviour and 
conform to shape constraints.

\subsection{Goodness-of-fit}\label{numerical-results}

Table \ref{tab:nmse-test} shows the median  
NMSE of best solutions over 30 runs obtained by each algorithm, with and without shape constraints,
on the test sets. The training results can be found in the supplementary material.

The table has $6$ quadrants: 
the left block shows the results without side information, the right block shows the results with shape constraints;
the top panel shows results for RW instances, 
the middle panel the results for the FDE and FYE instances, 
and the bottom panel the results for the same instances including $0.25$\% noise.

\begin{table}[!htbp]
	\caption{Median NMSE values for the test data. Values are multiplied by $100$ (percentage) and truncated at the second decimal place.}
	\label{tab:nmse-test}
\begin{tabular}[]{c@{}l>{$}r<{$}>{$}r<{$}>{$}r<{$}>{$}r<{$}|>{$}r<{$}>{$}r<{$}>{$}r<{$}>{$}r<{$}@{}}
\toprule
& & \multicolumn{4}{c|}{w/o. info} & \multicolumn{4}{c}{w. info}  \tabularnewline

& & \text{GP} & \text{GPC} & \text{ITEA} & \text{AML} & \text{GP} & \text{GPC} & \text{FIIT} & \text{SCPR} \tabularnewline

\hline
& Friction $\mu_{\operatorname{dyn}}$  & 8.28 & 7.73  & \textBF{4.35} & 12.99& 12.53 & 16.30& 35.90 & \textBF{8.07} \tabularnewline
& Friction $\mu_{\operatorname{stat}}$ & 7.22 & 5.44  & \textBF{4.46} & 6.82& 9.98 & 7.76 & 11.83 & \textBF{1.77} \tabularnewline
& Flow stress                          & 8.36 & 4.66  & --  & \textBF{0.15} & 34.05& 26.04 & 68.16 & \textBF{19.46} \tabularnewline
& Cars                                 & 75.18 & 76.23 & 75.06 & \textBF{74.72}& 76.86 & 77.67 & 76.64 & \textBF{73.83} \tabularnewline 
\hline
\parbox[t]{1.5em}{\multirow{19}{*}{\rotatebox[origin=c]{90}{without noise}}} 
& Aircraft lift                        &  0.63 & 0.15 & 0.22 & \textBF{0.00}&  0.80 &  1.01 & 0.14 & \textBF{0.00} \tabularnewline
& Flow psi                             &  0.75 & 0.13 & 0.05 & \textBF{0.00}&  4.80 &  5.36 & 2.91 & \textBF{0.00} \tabularnewline
& Fuel flow                            &  \textBF{0.00} & \textBF{0.00} & \textBF{0.00} & \textBF{0.00}&   \textBF{0.00} &  \textBF{0.00} & \textBF{0.00} & \textBF{0.00} \tabularnewline
& Jackson 2.11                         &  \textBF{0.00} & \textBF{0.00} & \textBF{0.00} & 0.62&  \textBF{0.00} &  \textBF{0.00} & \textBF{0.00} & 0.90 \tabularnewline
& Wave Power                           & 14.55 &30.82 & \textBF{2.26} & 2.74& 18.71 & 80.34 &21.31 & \textBF{13.50} \tabularnewline
& I.6.20                               &  0.46 & \textBF{0.00} & 0.31 & \textBF{0.00}&  1.61 & 0.42 & 3.20 & \textBF{0.01} \tabularnewline
& I.9.18                               &  2.88 & 2.49 & \textBF{0.91}& 4.98&  4.03 & 16.16 & \textBF{0.74} & 1.20 \tabularnewline
& I.15.3x                              &  0.34 & \textBF{0.01} & \textBF{0.01} & 0.02&  0.36 &  0.04 & \textBF{0.01} & \textBF{0.01} \tabularnewline
& I.15.3t                              &  0.21 & 0.01 & \textBF{0.00}& \textBF{0.00}&  0.15 &  0.03 & \textBF{0.00} & \textBF{0.00}  \tabularnewline
& I.30.5                               &  \textBF{0.00} & \textBF{0.00} & \textBF{0.00} &0.18 & \textBF{0.00} &  \textBF{0.00} & \textBF{0.00} & \textBF{0.00} \tabularnewline
& I.32.17                              &  \textBF{0.76} & 1.13 & 8.07 & 42.36&  \textBF{2.07} & 12.76 & 2.42 & 7.79 \tabularnewline
& I.41.16                              &  2.78 & 2.29 & \textBF{1.08} &15.14 &  8.99 & 17.72 & 5.15 & \textBF{1.56} \tabularnewline
& I.48.20                              &  \textBF{0.00} & \textBF{0.00} & \textBF{0.00} & \textBF{0.00}&  \textBF{0.00} &  \textBF{0.00} & \textBF{0.00} & \textBF{0.00} \tabularnewline
& II.6.15a                             &  3.55 & \textBF{2.50} & 4.66 & 16.17& 4.67 &  7.30 & 32.12 & \textBF{1.01} \tabularnewline
& II.11.27                             &  \textBF{0.00} & \textBF{0.00} & \textBF{0.00} & \textBF{0.00}&  \textBF{0.00} &  0.07 & \textBF{0.00} & \textBF{0.00} \tabularnewline
& II.11.28                             &  \textBF{0.00} & \textBF{0.00} & \textBF{0.00} & \textBF{0.00}&  \textBF{0.00} &  \textBF{0.00} & \textBF{0.00} & \textBF{0.00} \tabularnewline
& II.35.21                             &  3.29 & \textBF{1.18} & 2.61 & 1.40&  3.67 &  6.10 & 3.22 & \textBF{1.34} \tabularnewline
& III.9.52                             &116.25 &20.44 &66.16 & \textBF{19.88}&104.56 &106.48 & 71.93 & \textBF{33.41} \tabularnewline
& III.10.19                            &  0.41 & 0.04 & 0.17 & \textBF{0.01}&  0.55 &  0.33 & 0.31 & \textBF{0.00} \tabularnewline
\hline
\parbox[t]{1.5em}{\multirow{21}{*}{\rotatebox[origin=c]{90}{with noise}}} 
& Aircraft lift                             &  0.45 & 0.26 & \textBF{0.25} & 0.32 &  1.24 & 1.30 & \textBF{0.28} & 0.46\tabularnewline
& Flow psi                                  &  0.75 & 0.29 & \textBF{0.21} & 0.37& 5.90 & 6.02 & 4.65 & \textBF{0.57} \tabularnewline
& Fuel flow                                 &  0.21 & 0.24 & \textBF{0.18} & 0.34 & 0.30 & 0.30 & \textBF{0.25} &  \textBF{0.25} \tabularnewline
& Jackson 2.11                              &  \textBF{0.28} & 0.31 & 0.38 & 3.18& \textBF{0.24} & 0.25 & 0.30 & 0.83 \tabularnewline
& Wave Power                                & \textBF{21.23} &51.36 &99.88 &44.83 & 22.36 &68.96 & 21.39 & \textBF{11.88} \tabularnewline
& I.6.20                                    &  1.09 & \textBF{0.40} & 0.56 &0.45 & 2.14 & 0.78 & 3.61 & \textBF{0.55} \tabularnewline
& I.9.18                                    &  3.77 & 3.55 & \textBF{1.56} &4.02 & 5.25 &15.70 & \textBF{1.33} & 1.62 \tabularnewline
& I.15.3x                                   &  0.55 & \textBF{0.36} & 0.38 & 0.37& 0.56 & \textBF{0.35} & 0.36 & 0.42 \tabularnewline
& I.15.3t                                   &  0.65 & \textBF{0.48} & 0.58 &0.53 & 0.59 & 0.51 & 0.48 & \textBF{0.45} \tabularnewline
& I.30.5                                    &  \textBF{0.34} & 0.35 & 0.62 & 0.81 & \textBF{0.32} & 0.33 & 0.34 & 0.39 \tabularnewline
& I.32.17                                   &  \textBF{0.78} & 3.14 & 8.50 & 47.60& 3.95 &14.02 & \textBF{2.53} & 6.22 \tabularnewline
& I.41.16                                   &  3.13 & \textBF{2.32} & 3.47 &15.19 & 6.68 &19.72 & 5.05 & \textBF{2.93} \tabularnewline
& I.48.20                                   &  0.37 & 0.36 & 0.51 & \textBF{0.35}& \textBF{0.32} & \textBF{0.32} & 0.34 & \textBF{0.32} \tabularnewline
& II.6.15a                                  &  3.08 & \textBF{2.88} & 7.56 & 19.29& 3.87 & 6.05 &45.32 & \textBF{1.76} \tabularnewline
& II.11.27                                  &  \textBF{0.35} & 0.39 & 1.06 & 0.62& \textBF{0.37} & 0.61 & 0.41 & 0.47 \tabularnewline
& II.11.28                                  &  \textBF{0.38} & 0.44 & 0.39 & 0.51& \textBF{0.27} & 0.29 & 0.38 & 0.30 \tabularnewline
& II.35.21                                  &  3.88 & \textBF{1.33} & 2.43 & 2.10& 4.38 & 7.49 & 4.27 & \textBF{1.34} \tabularnewline
& III.9.52                                  &126.84 &\textBF{18.91} &74.08 & 24.81& 106.56 &90.18 & 73.44 & \textBF{32.69} \tabularnewline
& III.10.19                                 &  0.85 & \textBF{0.38} & 0.70 & 0.64& 0.91 & 0.64 & 0.70 & \textBF{0.46} \tabularnewline
\bottomrule
\end{tabular}
\end{table}

Analysing the detailed results we observe that the best result of
all models with information is better for 18 of 42 instances (RW: 2, no noise: 6, noisy: 10). While the
best result of models without information is better for 14 of 42
instances (RW: 2, no noise: 3, noisy: 9).

Within both groups there are algorithms with significantly 
different test errors (without info: p-value: $0.0103$, with info:
p-value: $6.467\cdot 10 ^{-6}$, Friedman's rank sum test with Davenport correction).
Pairwise comparison of results without info shows that GPC is better
than GP (p-value: 0.011) and AML (p-value: 0.043) using Friedman's
rank sum test with Bergman correction. For the problem instances with noise
AML produced the best result for only 1 out of 19 instances. For the instances
without noise the results of AML are similar to results of GP, GPC, and ITEA.
Pairwise comparison of results with info shows that SCPR is better than the
other algorithms and no statistically significant difference was found between GP, GPC and FIIT.
The p-values for all pairwise tests are shown in Table \ref{tab:statistics}.

Comparing the results with and without constraints for each algorithm
individually, we find that the results are in general worse when using
constraints. GP is better than GP (info) for 27 instances. GPC is
better than GPC (info) for 32 instances. ITEA is better than FIIT for
19 instances.
For the RW data sets, ITEA managed to find the best  
models on two out of the four instances. 
For \emph{Flow Stress}, 
ITEA returned a solution that produced numerical errors
for the test set. 
This is not the case when we include the shape-constraints, 
as we can see on the top-right quadrant. 
In this situation, FI-2POP-IT was capable of finding expressions that 
did not return invalid results for the test set. 

\begin{table}
  \centering
\begin{tabular}{llll}
  \multicolumn{4}{c}{without info} \\
  GP & GPC & ITEA & AML \\
  \hline
n/a & \textBF{0.011} & 0.325 & 0.499 \\
\textBF{0.011} & n/a & 0.325 & \textBF{0.043} \\
0.325 & 0.325 & n/a & 0.353 \\
0.499 & \textBF{0.043} & 0.353 & n/a \\
\end{tabular}
\begin{tabular}{llll}
  \multicolumn{4}{c}{with info} \\
  GP & GPC & FIIT & SCPR \\
  \hline
n/a & 0.151 & 0.353 & \textBF{0.002} \\
0.151 & n/a & 0.054 & \textBF{0.000} \\
0.353 & 0.054 & n/a & \textBF{0.028} \\
\textBF{0.002} & \textBF{0.000} & \textBF{0.028} & n/a \\
\end{tabular}
\caption{\label{tab:statistics}p-values for pairwise comparison of algorithms in both groups  (Friedman's rank sum test with Bergman correction)}
\end{table}

\subsection{Computational Overhead}

Another important impact of introducing constraints that should be
considered is the computational overhead introduced to each approach.

For ITEA the execution time is approximately doubled when using
 the constraint handling. The reason being that
parameter optimization is much easier for the ITEA representation and
the calculation of the partial derivatives is a simple mechanical
process as shown in
Section~\ref{defining-constraints-with-interval-arithmetic}.  For GP and GPC
the execution time factor is approximately $5$ when including shape constraints. The increased execution time for
GP results from the additional effort for building the partial
derivatives for each solution candidate and for the interval
evaluation.  
We observed that the increase in execution time is less extreme for
problem instances with a large number of rows where the relative
effort for symbolic derivation of solution candidates becomes smaller.

\section{Discussion}\label{discussion}

The results presented on the previous section largely corroborate our initial
assumptions for shape-constrained symbolic regression. First of all,
when we do not explicitly consider shape constraints within SR
algorithms we are unlikely to find solutions which conform to expected behaviour. 
We showed that the results produced by the two newly-introduced algorithms 
in fact conform to shape constraints.
Our assessment of the extrapolation and interpolation behaviour of SR
models highlighted the bad extrapolation behaviour as well as
occasional problems even for interpolation. The improved results when
including shape constraints support the argument to include interval
arithmetic to improve the robustness of SR
solutions~\citep{keijzer2003improving,pennachin2010robust}.

However, our results also show that including shape-constraints via interval arithmetic leads
to SR solutions with higher prediction errors on training and test
sets. While the increased error on the training set is expected, we
hoped we would be able to improve prediction errors on the test set.
Assuming that the constraints are
correct, this should hypothetically be possible because the shape
constraints provide additional information for improving the fit on
noisy data sets.
In fact, we observed that the results got worse when using information for 
tree-based GP with and without local optimization (GPC).
There are several possible explanations such as slower convergence, more rapid
loss of diversity, or rejection of feasible solutions because of the
pessimistic bounds produced by interval
arithmetic. Another hypothesis is that the positive effect of shape constraints
becomes more relevant with higher noise levels. 
We are not able to give a conclusive answer for the main 
cause of the higher prediction errors with side information and leave this question open for 
future research.

Comparison with AutoML as implemented by auto-sklearn showed that GP with parameter optimization (GPC) 
produced better test results than
AutoML (p-value: $0.011$) over the benchmark set without
shape-constraints.  However, AutoML does not support monotonicity constraints and, because of that, 
we cannot use it to compare with the results using side information. Therefore, 
we compared the results of our proposed algorithms with shape-constrained polynomial regression (SCPR). 
The results show that SCPR performs better than the evolutionary algorithms for this benchmark set, 
which indicates that we can find a good approximation of many of our benchmark instances using polynomials. 

An advantage of SCPR is that it is formulated as a convex optimization
problem that can be solved efficiently and deterministically by
highly-tuned solvers. A drawback of SCPR is the potentially large
model size. The number of terms of a homogeneous $n$-variate polynomial 
of degree $d$ is $\binom{n+d}{n}$. Our experiments found that polynomial 
degrees of up to eight were required to find a good fit. The studied problems had five 
variables on average, which led to more than a thousand terms in our polynomial models. 
The models produced by the evolutionary algorithms were, on average, much smaller as we used a 
limit of $50$ nodes for GP expression trees and $6$ terms for ITEA/FI-2POP-IT.

\section{Conclusions}\label{conclusions}

In this paper we have introduced shape-constrained symbolic regression
which allows to include prior knowledge into
SR. Shape-constrained symbolic regression allows to enforce that the
model output must be within given bounds, or that outputs must be
monotonically increasing or decreasing over selected inputs. The
structure and the parameters of the symbolic regression model are
however still identified by the algorithm.

We have described two algorithms for shape-constrained symbolic
regression which are extensions of tree-based genetic programming with
optional local optimization and the Interaction-Transformation
Evolutionary Algorithm that uses a more restrictive representation
with the goal of returning simpler expressions.

The extensions add the ability to calculate partial derivatives for any
expression generated by the algorithms and use interval arithmetic to determine
bounds and determine if models conform to shape constrains. Two approaches for
handling constraint violations have been tested. The first approach simply
adjusts fitness for infeasible solutions while the second approach splits the
population into feasible and infeasible solutions.

The results showed the importance of treating the shape constraints
inside the algorithms. First of all, we have collected more evidence that without any
feasibility control, is unlikely to find feasible solutions for
most of the problems. Following, we verified the efficacy of our
approach by measuring the frequency of infeasible solutions and
reporting the median numerical error of our models. The modified
algorithms were all capable of finding models conforming to the shape
constraints. This shows that the introduction of shape
constraints can help us finding more realistic models.
However, we have also found that the extended algorithms with
shape constraints produce worse solutions on the test set on average. 

For the next steps we intend to analyse in detail the causes for the
worse solutions with shape-constrained SR. The bounds determined via interval arithmetic are
very wide and might lead to rejection of feasible solutions as well as
premature convergence.
This is an issue that could potentially be
solved by using more elaborate bound estimation schemes such as affine arithmetic or recursive splitting.
Other possibilities for the
constraint-handling include multi-objective optimization and penalty functions.
Alternatively, optimistic approaches (e.g. using sampling) or a hybridization of pessimistic and optimistic approaches for shape-constrained regression can be used to potentially improve the results.
Additionally, it would be worthwhile to study the effects of constraint-handling mechanisms on population diversity in more detail.

\section{Acknowledgments}\label{acknowledgments}

This project is partially funded by Funda\c{c}\~{a}o de Amparo \`{a}
Pesquisa do Estado de S\~{a}o Paulo (FAPESP), grant number
2018/14173-8. And some of the experiments (ITEA) made use of the
Intel\textregistered AI DevCloud, which Intel\textregistered provided
free access.

The authors gratefully acknowledge support by the Christian Doppler
Research Association and the Federal Ministry of Digital and Economic
Affairs within the Josef Ressel Centre for Symbolic Regression.

\bibliographystyle{apalike}
\bibliography{manuscript}

\section{Supplementary}
\subsection{Problem definition details}
Table \ref{tab:problem-instances-constraints} shows the input space
and the monotonicity constraints that we have used for the problem
instances.
\begin{table}
  \footnotesize
  \begin{tabular}{p{1.75cm}>{$}c<{$}>{$}c<{$}}
    Name & \text{Input space} & \text{Constraints} \tabularnewline
    \hline
    Aircraft lift & (C_{L\alpha},\alpha, C_{L\delta_e}, \delta_e , S_{\operatorname{HT}}, S_{\operatorname{ref}}) & (1,1,1,1,1,-1) \tabularnewline
                  & \in [0.3..0.9] \times [2..12] \times [0.3..0.9] \times [0..12] \times [0.5..2] \times [3..10]& \tabularnewline
    Flow psi & (V_\infty, R, \Gamma, r, \theta) & (1, 1, 1, -1, 1)\tabularnewline
             & \in [30..100] \times [0.1..0.5] \times [2..15] \times [0.5..1.5] \times [10..90]  & \tabularnewline
    Fuel flow & (A\star, p_0, T_0) & (1, 1, -1)\tabularnewline
              & \in [0.2..2]\times[3\cdot10^5..7\cdot10^5]\times[200..400] & \tabularnewline
    Jackson 2.11 & \left( q,y,\mathit{Volt},d,\epsilon\right) & (1, -1, 1, 1, 1)\tabularnewline
                 & \in [1..5]\times[1..3]\times[1..5]\times[4..6]\times[1..5] & \tabularnewline
    Wave power & \left( G,c,\mathit{m1},\mathit{m2},r\right) & (-1, 1, -1, -1, 1)\tabularnewline
               & \in [1..2]\times[1..2]\times[1..5]\times[1..5]\times[1..2] & \tabularnewline
    I.6.20 & (\sigma, \theta) \in [1..3]^2 & (0, -1)\tabularnewline
    I.9.18 & \left( \mathit{m1},\mathit{m2},G,\mathit{x1},\mathit{x2},\mathit{y1},\mathit{y2},\mathit{z1},\mathit{z2}\right)  & (1, 1, 1, -1, 1, \tabularnewline
    &  \in [1..2]\times[1..2]\times[1..2]\times[3..4] &  -1, 1, -1, 1) \tabularnewline
    & \times[1..2]\times[3..4]\times[1..2]\times[3..4]\times[1..2] & \tabularnewline
    I.15.3x & (x, u, c, t) \in [5..10]\times[1..2]\times[3..20]\times[1..2] & (1, 0, -1, -1)\tabularnewline
    I.15.3t & (x,c,u,t) \in [1..5]\times[3..10]\times[1..2]\times[1..5]  & (0, 0, 0, 1) \tabularnewline
    I.30.5 & \left( \mathit{lambd},d,n\right) \in [1..5]\times[2..5]\times[1..5] & (1, -1, -1)\tabularnewline
    I.32.17 & \left( \epsilon,c,\mathit{Ef},r,\omega,{{\omega}_0}\right) & (1, 1, 1, 1, 1, -1) \tabularnewline
            & \in [1..2]\times[1..2]\times[1..2]\times[1..2]\times[1..2]\times[3..5] & \tabularnewline
    I.41.16 & \left( \omega,T,h,\mathit{kb},c\right)  \in [1..5]^5 & (0, 1, -1, 1, -1) \tabularnewline
    I.48.20 & (m, v, c) \in [1..5]\times[1..2]\times[3..20] & (1, 1, 1)\tabularnewline
    II.6.15a & \left( \epsilon,{p_d},r,x,y,z\right) \in [1..3]^6  & (-1, 1, -1, 1, 1, 1)\tabularnewline
    II.11.27 & \left( n,\alpha,\epsilon,\mathit{Ef}\right) \in [0..1]\times[0..1]\times[1..2]\times[1..2]   & (1, 1, 1, 1) \tabularnewline
    II.11.28 & \left( n,alpha\right) \in [0..1]^2 & (1, 1) \tabularnewline
    II.35.21 & \left( {n_{\mathit{rho}}},\mathit{mom},B,\mathit{kb},T\right)  \in [1..5]^5  & (1, 1, 1, -1, -1) \tabularnewline
    III.9.52 & \left( {p_d},\mathit{Ef},t,h,\omega,{{\omega}_0}\right) & (1, 1, 0, -1, 0, 0) \tabularnewline
             & \in [1..3]\times[1..3]\times[1..3]\times[1..3]\times[1..5]\times[1..5] & \tabularnewline
    III.10.19 & \left( \mathit{mom},\mathit{Bx},\mathit{By},\mathit{Bz}\right)  \in [1..5]^4  & (1, 1, 1, 1) \tabularnewline
    & & \tabularnewline
    \hline
    Friction $\mu_{\operatorname{dyn}}$ & (p, v, T) \in [0.1..15]\times[0.01..3]\times[-50..250] & (-1, -1, -1) \tabularnewline
    Friction $\mu_{\operatorname{stat}}$ &  (p, v, T) \in  [0.1..15]\times[0.01..3]\times[-50..250]  & (-1, 0, -1)\tabularnewline
    Flow stress &  (\phi, \dot\phi, T) \in  [0..1]\times[0.001..10]\times[250..600] & (0, 1, -1) \tabularnewline
    Cars & (\mathit{cyl}, \mathit{dis}, \mathit{hp}, w, \mathit{acc}) & (0, -1, -1, -1, 0) \tabularnewline
         & \in [3..8]\times [68..455]\times [46..230]\times [8..24.8] & \tabularnewline
  \end{tabular}
  \caption{\label{tab:problem-instances-constraints}Input space and the monotonicity constraints we used for all benchmark instances. In the constraints tuple a negative value for $c_i$ means that the model is non-increasing over the $i$-th variable, a positive value means that the model is non-decreasing and a zero means that there is no monotonicity constraint.}
\end{table}

\subsection{Training results}

    Table \ref{tab:nmse-train} shows the median NMSE values (in percent) on the training set for the best solutions over 30 runs of each algorithm. The test results are given in the paper.

\begin{table}[!htbp]
	\caption{Median of the NMSE for the training data without and with shape constraints. Values are multiplied by $100$ (percentage) and truncated at the second decimal place.}
	\label{tab:nmse-train}
\begin{tabular}[]{c@{}l>{$}r<{$}>{$}r<{$}>{$}r<{$}>{$}r<{$}|>{$}r<{$}>{$}r<{$}>{$}r<{$}>{$}r<{$}@{}}
\toprule
& & \multicolumn{4}{c|}{w/o. info} & \multicolumn{4}{c}{w. info}  \tabularnewline

& & \text{GP} & \text{GPC} & \text{ITEA} & \text{AML} & \text{GP} & \text{GPC} & \text{FIIT} & \text{SCPR} \tabularnewline

\hline

& Friction $\mu_{\operatorname{dyn}}$  & 0.80 & 0.48 & 0.24 & \textBF{0.07}  & 2.02 & \textBF{1.51} & 1.98 & 3.05 \tabularnewline
& Friction $\mu_{\operatorname{stat}}$ & 0.41 & 0.30 & 0.25 & \textBF{0.11} & 1.84 & 0.95 & 5.81 & \textBF{0.61} \tabularnewline
& Flow stress                          & 4.73 & 4.85 & 4.70 & \textBF{0.05}  & 23.57& 22.76& 42.55 & \textBF{14.27} \tabularnewline
& Cars                                 & 9.34 & 10.78& 7.48 & \textBF{6.42} & 10.79& 11.26& 11.56 & \textBF{9.78} \tabularnewline
\hline
\parbox[t]{1.5em}{\multirow{19}{*}{\rotatebox[origin=c]{90}{without noise}}} 
& Aircraft lift                        & 0.33 & 0.08 & 0.08 & \textBF{0.00}  & 0.49 & 0.68 & 0.13 & \textBF{0.00} \tabularnewline
& Flow psi                             & 0.54 & 0.14 & 0.02 &  \textBF{0.00} & 4.54 & 4.59 & 3.63 & \textBF{0.00} \tabularnewline
& Fuel flow                            & \textBF{0.00} & \textBF{0.00} & \textBF{0.00} & \textBF{0.00}  & \textBF{0.00} & \textBF{0.00} & \textBF{0.00} & \textBF{0.00} \tabularnewline
& Jackson 2.11                         & \textBF{0.00} & \textBF{0.00} & \textBF{0.00} & 0.10  & \textBF{0.00} & \textBF{0.00} & \textBF{0.00} & \textBF{0.00} \tabularnewline
& Wave Power                           & 0.34 & 0.69 & \textBF{0.01} & 0.41 & 1.07 & 6.88 & 0.93 & \textBF{0.00} \tabularnewline
& I.6.20                               & 0.18 & \textBF{0.00} & 0.01 & \textBF{0.00}  & 1.47 & 0.34 & 3.61 & \textBF{0.00} \tabularnewline
& I.9.18                               & 0.94 & 0.63 & \textBF{0.17} &  0.47 & 1.15 & 4.48 & 0.18 & \textBF{0.00} \tabularnewline
& I.15.3x                              & 0.18 & \textBF{0.00} & \textBF{0.00} & \textBF{0.00}  & 0.19 & 0.01 & \textBF{0.00} & \textBF{0.00} \tabularnewline
& I.15.3t                              & 0.11 & \textBF{0.00} & \textBF{0.00} & \textBF{0.00}  & 0.10 & 0.01 & \textBF{0.00} & \textBF{0.00} \tabularnewline
& I.30.5                               & \textBF{0.00} & \textBF{0.00} & \textBF{0.00} & 0.22 & \textBF{0.00} & \textBF{0.00} & \textBF{0.00} & \textBF{0.00} \tabularnewline
& I.32.17                              & 0.18 & 0.13 & \textBF{0.05} & 20.19 & 0.43 & 4.89 & \textBF{0.06} & 0.11 \tabularnewline
& I.41.16                              & 0.19 & 0.18 & \textBF{0.03} & 12.11 & 1.13 & 1.52 & 0.24 & \textBF{0.00} \tabularnewline
& I.48.20                              & \textBF{0.00} & \textBF{0.00} & \textBF{0.00} & \textBF{0.00} & \textBF{0.00} & \textBF{0.00} & \textBF{0.00} & \textBF{0.00} \tabularnewline
& II.6.15a                             & 0.64 & \textBF{0.15} & 0.18 & 19.98 & 1.16 & 1.24 & 0.61 & \textBF{0.00} \tabularnewline
& II.11.27                             & \textBF{0.00} & \textBF{0.00} & \textBF{0.00} & \textBF{0.00} & \textBF{0.00} & 0.04 & \textBF{0.00} & \textBF{0.00} \tabularnewline
& II.11.28                             & \textBF{0.00} & \textBF{0.00} & \textBF{0.00} & \textBF{0.00} & \textBF{0.00} & \textBF{0.00} & \textBF{0.00} & \textBF{0.00} \tabularnewline
& II.35.21                             & 1.75 & 0.49 & 0.44 & \textBF{0.17} & 2.43 & 3.95 & 2.08 & \textBF{0.00} \tabularnewline
& III.9.52                             &23.69 & 7.14 & 8.13 & \textBF{5.44} &32.27 &32.56 & 21.45 & \textBF{0.20} \tabularnewline
& III.10.19                            & 0.31 & 0.02 & 0.08 & \textBF{0.00} & 0.37 & 0.23 & 0.21 & \textBF{0.00} \tabularnewline
\hline
\parbox[t]{1.5em}{\multirow{20}{*}{\rotatebox[origin=c]{90}{with noise}}} 
& Aircraft lift                             & 0.45 & 0.26 & 0.25 & \textBF{0.22}  & 0.76 & 0.87 & \textBF{0.28} & 3.94 \tabularnewline
& Flow psi                                  & 0.75 & 0.29 & \textBF{0.21} & \textBF{0.21}  & 5.69 & 5.22 & \textBF{4.65} & 4.78 \tabularnewline
& Fuel flow                                 & 0.21 & 0.24 & \textBF{0.18} & 0.24 & 0.26 & 0.27 & \textBF{0.25} & 5.32 \tabularnewline
& Jackson 2.11                              & 0.28 & 0.29 & \textBF{0.20} & 0.66 & 0.32 & 0.34 & 0.31 & \textBF{0.09}  \tabularnewline
& Wave Power                                & \textBF{0.52} & 1.12 & 7.24 & 2.42 & 0.92 & 4.74 & 0.94 & \textBF{0.07} \tabularnewline
& I.6.20                                    & 0.56 & \textBF{0.23} & 0.25 & 0.43 & 1.79 & 0.77 & 3.86 & \textBF{0.21} \tabularnewline
& I.9.18                                    & 1.15 & 0.87 & \textBF{0.25} & 0.46 & 1.47 & 4.74 & 0.26 & \textBF{0.12} \tabularnewline
& I.15.3x                                   & 0.35 & 0.16 & \textBF{0.15} & 0.28 & 0.36 & 0.19 & \textBF{0.15} & 0.18 \tabularnewline
& I.15.3t                                   & 0.34 & 0.23 & \textBF{0.19} & 0.25 & \textBF{0.15} & 0.26 & 0.21 & 0.21 \tabularnewline
& I.30.5                                    & 0.20 & 0.19 & \textBF{0.15} & 0.76 & 0.21 & 0.21 & 0.20 & \textBF{0.14} \tabularnewline
& I.32.17                                   & 0.85 & 0.81 & \textBF{0.65} &28.23 & 1.74 & 6.36 & 0.71 & \textBF{0.11} \tabularnewline
& I.41.16                                   & 0.40 & 0.30 & \textBF{0.16} &13.37 & 1.03 & 2.10 & 0.49 & \textBF{0.05}  \tabularnewline
& I.48.20                                   & 0.16 & 0.16 & \textBF{0.14} & 0.23 & 0.20 & 0.21 & \textBF{0.19} & \textBF{0.19} \tabularnewline
& II.6.15a                                  & 0.58 & 0.33 & \textBF{0.30} &20.72 & 1.03 & 1.30 & 0.61 & \textBF{0.00} \tabularnewline
& II.11.27                                  & 0.18 & 0.17 & \textBF{0.15} & 0.19 & 0.21 & 0.38 & \textBF{0.17} & 0.18 \tabularnewline
& II.11.28                                  & 0.30 & 0.29 & \textBF{0.24} & 0.38 & 0.36 & 0.35 & \textBF{0.33} & 0.35 \tabularnewline
& II.35.21                                  & 2.21 & 0.62 & \textBF{0.55} & \textBF{0.55} & 3.22 & 4.52 & 2.37 & \textBF{0.20} \tabularnewline
& III.9.52                                  & 29.77& 5.65 & 9.22 & \textBF{5.50} &33.68& 29.77 &22.17 & \textBF{0.28} \tabularnewline
& III.10.19                                 & 0.49 & 0.21 & 0.32 & \textBF{0.17} & 0.68 & 0.48 & 0.45 & \textBF{0.16} \tabularnewline
\bottomrule
\end{tabular}
\end{table}

\subsection{Sensitivity to population size}
For the experiments in the paper we have used parameter settings based on prior knowledge 
about settings which tend to produce good results on a wide range of problems. For instance
we have set the population size to 1000 individuals for GP and 200 for ITEA and FIIT. Similarly, we have chosen other parameter values.
It should be noted that we have not tried to tune the hyper-parameters (globally or for individual instances). 
Since population size is potentially a sensitive parameter it is worthwhile to analyse 
how the algorithm convergence is influenced by different parameter settings.
On the one hand smaller population sizes can improve the runtime on the other hand there is a danger that the population size becomes too small
leading to loss of diversity and premature convergence. 

\begin{figure}
   \centering
    \includegraphics[width=0.45\textwidth,trim={30mm 25mm 33mm 40mm},clip]{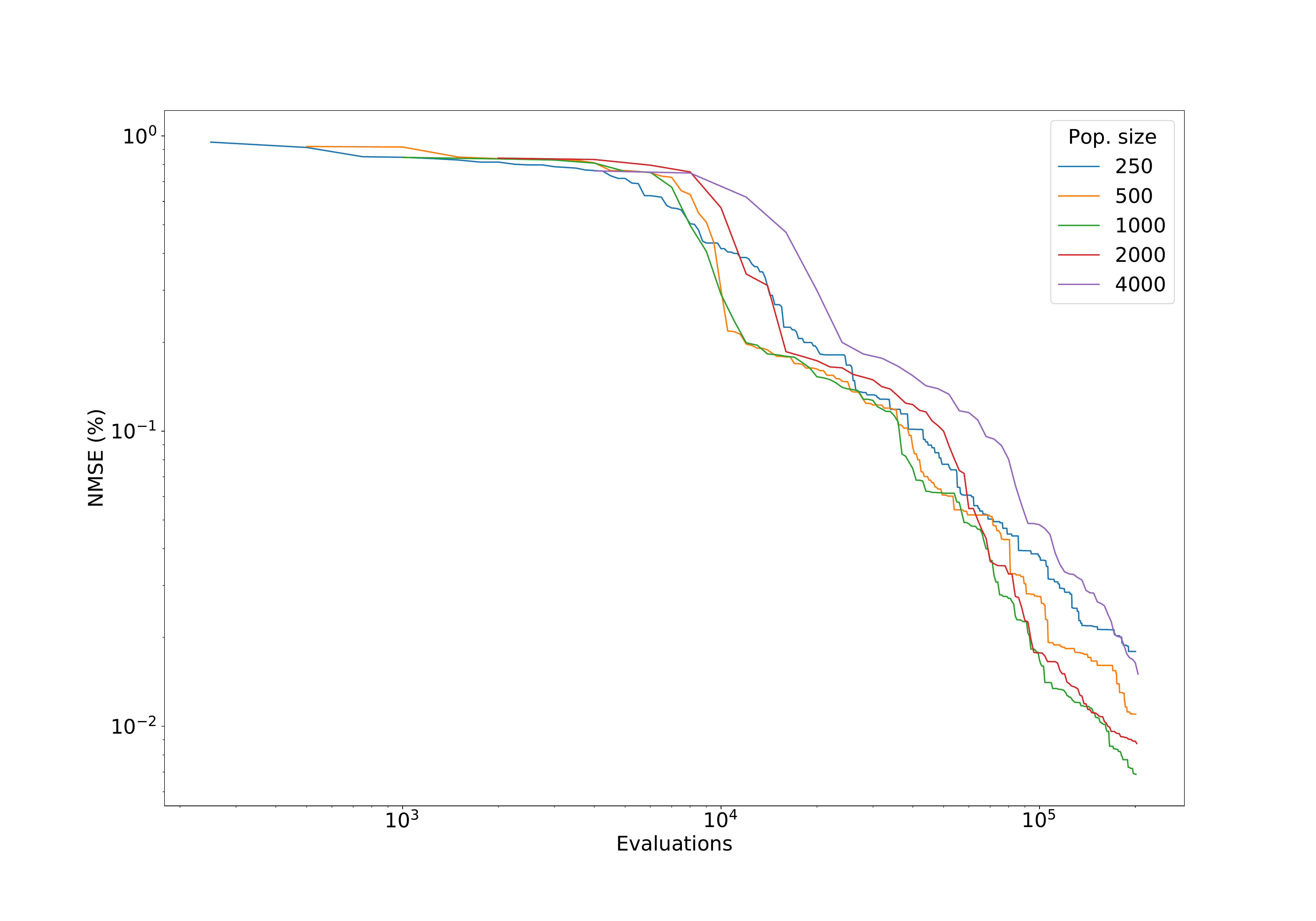}
	\includegraphics[width=0.45\textwidth,trim={30mm 25mm 33mm 40mm},clip]{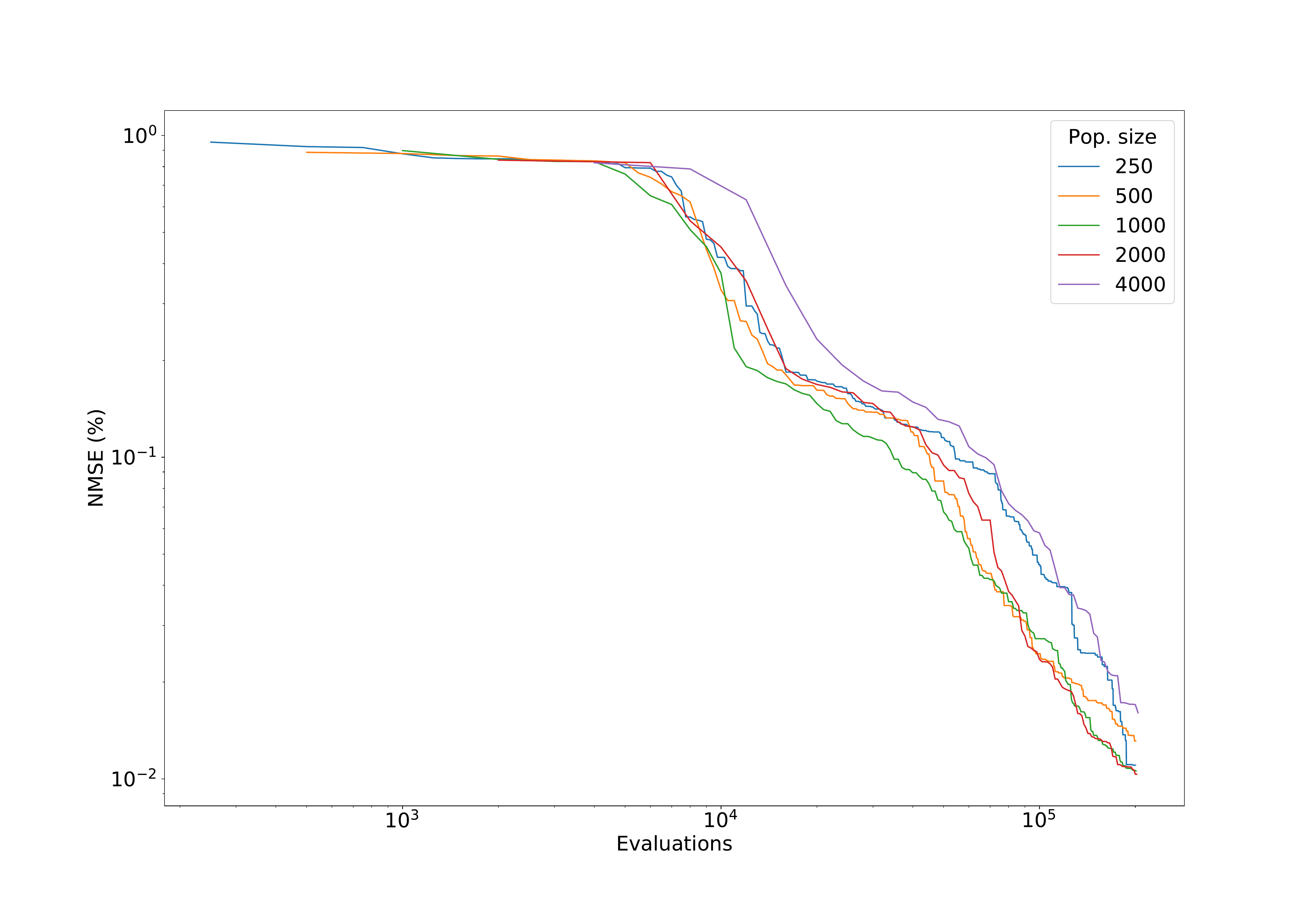}

	\includegraphics[width=0.45\textwidth,trim={30mm 25mm 33mm 40mm},clip]{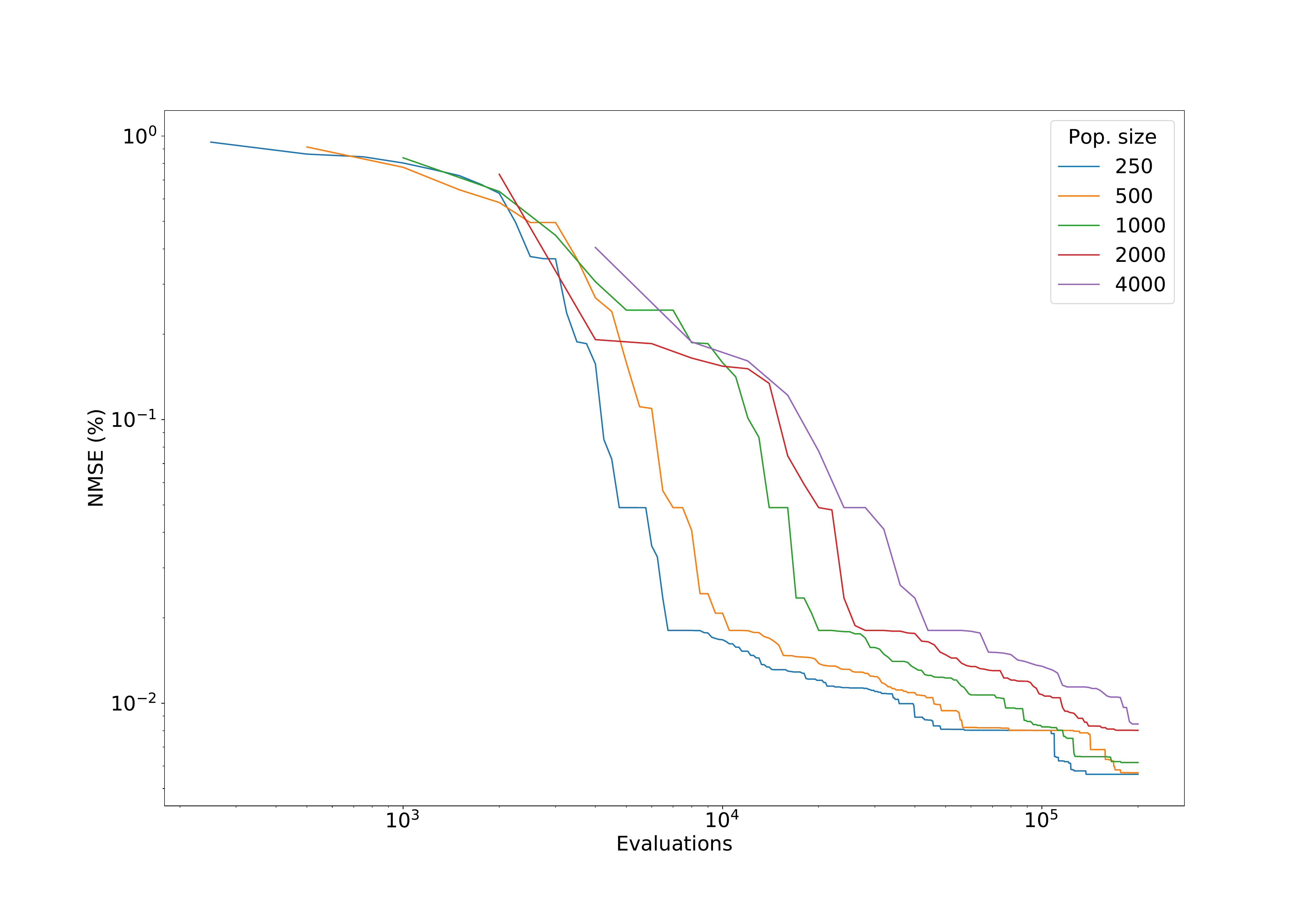}
	\includegraphics[width=0.45\textwidth,trim={30mm 25mm 33mm 40mm},clip]{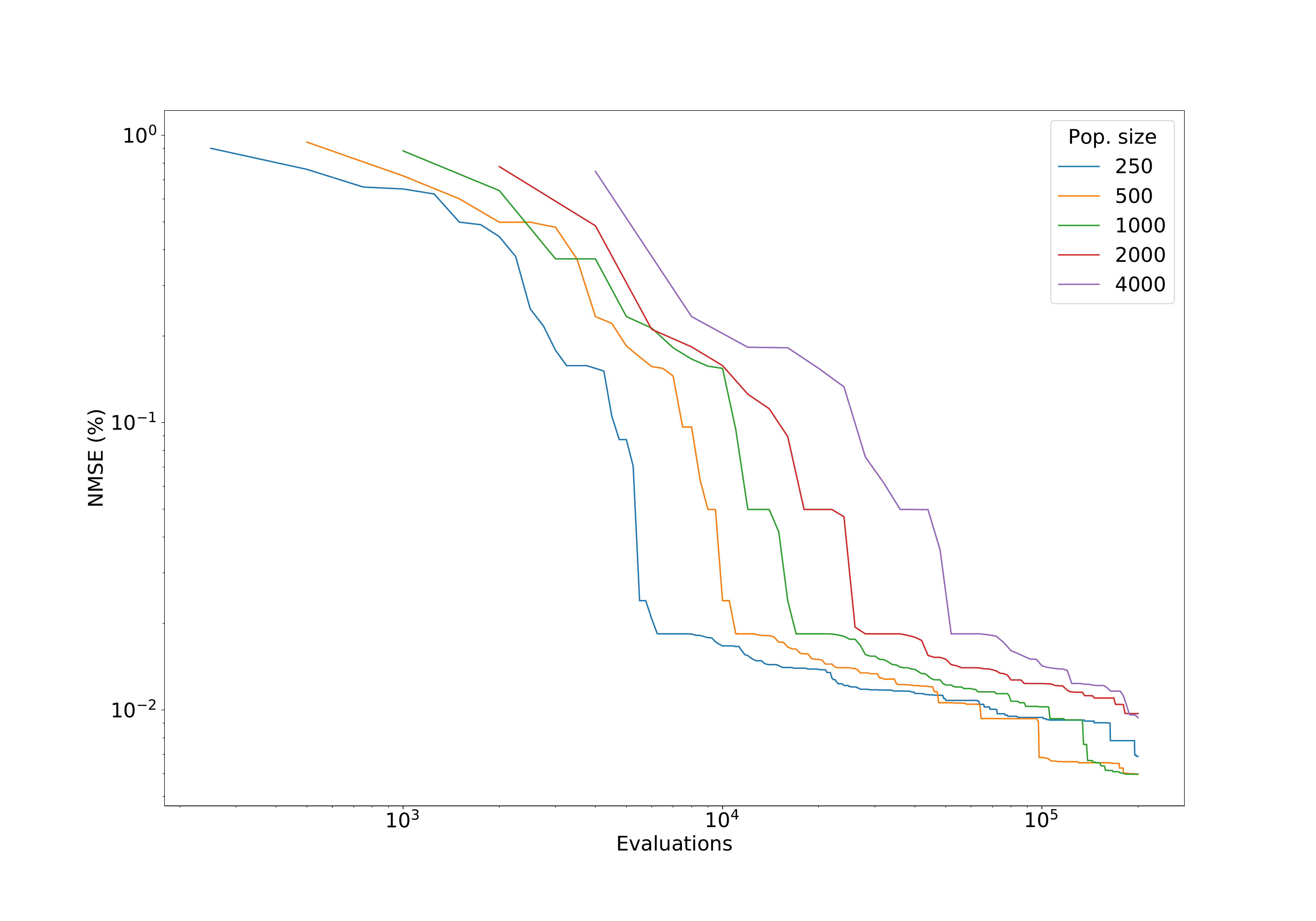}
	\caption{Convergence graphs of best quality over evaluations for different population sizes (top: GP, bottom: FI2POP, left: without noise, right with noise).}\label{fig:wave-pop-sensitivity}
\end{figure}

For the experiment we have chosen one of the harder problem instances (Wave power) and have used the population sizes: $250$, $500$, $1000$, $2000$, and $4000$.
Figure \ref{fig:wave-pop-sensitivity} shows the best error over the number of evaluations (note: log-log scale) for GP and FI2POP with and without noise. 
It can be seen that comparable results can also be achieved with smaller population sizes. Overall, the population size of 1000 individuals, that 
we have originally used for the GP experiments, performs well compared to the other population sizes. The plots show no overly long plateaus in quality which would be indicative
of premature convergence.

\end{document}